\documentclass{aidas}
\pdfoutput=1
\usepackage[toc,page,header]{appendix}

\usepackage{natbib}
\usepackage{latexsym}

\usepackage{url}
\usepackage{amssymb}
\usepackage[utf8]{inputenc}
\usepackage{booktabs}
\usepackage{pifont} 
\usepackage{multirow}
\usepackage{makecell}
\usepackage{paralist}
\usepackage{xspace}
\usepackage{color}
\usepackage{xcolor}
\usepackage{colortbl}
\usepackage{adjustbox}
\usepackage[edges]{forest}
\usepackage{tikz} 
\usepackage{wrapfig}
\usepackage{environ}
\usepackage{multicol}
\usepackage{cleveref} 
\usepackage{booktabs}
\usepackage{tabularx}
\usepackage{xcolor}
\usepackage{lipsum} 
\usepackage{amsmath,amssymb}

\usepackage{caption}
\usepackage{amsfonts}
\usepackage{caption}
\usepackage{placeins}
\usepackage{amssymb}
\usepackage{pifont}
\usepackage{bbding}
\usepackage{float}
\usepackage{afterpage}

\usepackage{fontawesome5} 
\usepackage{hyperref} 
\usepackage{enumitem}

\definecolor{lightblue}{RGB}{220,235,250}

\hypersetup{
    colorlinks,
    linkcolor={blue!80!black},
    citecolor={blue!80!black},
}
\tikzset{
    root/.style =             {align=center, text width=1cm, rounded corners=3pt, line width=0.3mm, fill=gray!10, draw=gray!80, font=\small},
    demographic/.style =         {align=center, text width=1.8cm, rounded corners=3pt, line width=0.3mm, fill=blue!10, draw=blue!80, font=\footnotesize},
    demographic_work/.style =    {align=center, text width=10cm, rounded corners=3pt, line width=0.3mm, fill=blue!10, draw=blue!0, font=\footnotesize},
    character/.style =         {align=center, text width=1.8cm, rounded corners=3pt, line width=0.3mm, fill=red!10, draw=red!80, font=\footnotesize},
    character_work/.style =    {align=center, text width=10cm, rounded corners=3pt, line width=0.3mm, fill=red!10, draw=red!0, font=\footnotesize},
    personalization/.style =           {align=center, text width=1.8cm, rounded corners=3pt, line width=0.3mm, fill=cyan!10, draw=cyan!80, font=\footnotesize},
    personalization_work/.style =      {align=center, text width=10cm, rounded corners=3pt, line width=0.3mm, fill=cyan!10, draw=cyan!0, font=\footnotesize},
    risk/.style =         {align=center, text width=1.8cm, rounded corners=3pt, line width=0.3mm, fill=orange!10, draw=orange!80, font=\footnotesize},
    risk_work/.style =    {align=center, text width=10cm, rounded corners=3pt, line width=0.3mm, fill=orange!10, draw=orange!0, font=\footnotesize},
}

\newcommand{\method}[1]{\textcolor{black}{{Dynin-Robotics}}{{#1}}}

\definecolor{darkgreen}{RGB}{0,128,0}

\newcommand{\cmark}{\ding{51}}
\newcommand{\xmark}{\ding{55}}

\usepackage{CJK}
\makeatletter
\@ifpackagelater{microtype}{2025/05/05}{}{%
  \DeclareRobustCommand\showhyphens[1]{%
    \setbox0\vbox{%
      \color@begingroup
      \everypar{}\parfillskip\z@skip
      \hsize\maxdimen\normalfont
      \pretolerance\m@ne\tolerance\m@ne
      \hbadness\z@\showboxdepth\z@\ #1%
      \color@endgroup
    }%
  }%
}
\makeatother

\title{\method{}: Omnimodal Unified Diffusion Vision-Language-Action Model}

\author{
Hoeun Lee$^{\S \P}$, 
Jaeik Kim$^{\P}$,
Jusang Oh$^{\P}$,
Jinhyeok Kim,
Geon Choi,
Hyeonggeun Kim,
Jaeyoung Do$^{\dagger}$
}

\affiliation{
    AIDAS Lab \\ Seoul National University \\[1.5ex]
    {\small
        \href{https://dynin.ai/robotics/}{\faHome~Project Page} \quad
        \href{https://github.com/AIDASLab/Dynin-Robotics}{\faGithub~Code} \quad
        \href{https://huggingface.co/snu-aidas/Dynin-Robotics}{\faDatabase~Model}
    }
    \vspace{-1.5em}
}

\abstract{\begin{abstract}
-Visual goal and dynamics prediction can provide language-conditioned robot policies with both a target outcome and a representation of action-dependent scene changes. We bring these predictions into action generation and selection through a shared trajectory model. Dynin-Robotics implements this formulation on Dynin-Omni, an omnimodal masked-diffusion backbone, representing language, visual observations, goals, and actions as discrete tokens. By varying the conditioning and target spans, the same model learns action prediction, action-conditioned next-observation prediction, terminal goal-state prediction, and trajectory-to-instruction reconstruction. These conditional interfaces support test-time scaling by allocating additional computation to goal prediction and action-candidate evaluation, as well as joint refinement of action and future-state predictions. We continually pretrain the model on approximately 1.33 million trajectories from 48 Open X-Embodiment datasets and adapt it separately to downstream domains. On two VLABench tasks, robot pretraining improves adaptation within a fixed Stage-2 step budget, and the full objective mixture improves shifted-instruction success over Policy-only post-training under the same coupled decoder. Combining goal guidance with joint action--next-state denoising further improves shifted-instruction success over action-only decoding; the benefit depends on how the predictions are composed. Dynin-Robotics achieves competitive performance on LIBERO and zero-shot LIBERO-Plus, together with a 78.4\% average success rate across four manipulation conditions on a Franka Research~3 robot. An optimized block-parallel implementation accelerates model-side action decoding by up to $29.2\times$ relative to the base implementation under the reported profiling setup. These results support shared trajectory modeling as a common interface for learning complementary robot objectives and composing their predictions during control.
\end{abstract}}

\begin{document}

\maketitle

\begingroup
\renewcommand\thefootnote{\fnsymbol{footnote}}
\footnotetext[1]{
$^{\S}$ Project lead. 
$^{\P}$ Core contributors. 
% $^{\ddagger}$ Supervision. 
$^{\dagger}$ Corresponding author.
}
\endgroup

\vspace{-20px}
\begin{figure}[ht!]
\centering
\includegraphics[width=1.0\linewidth]{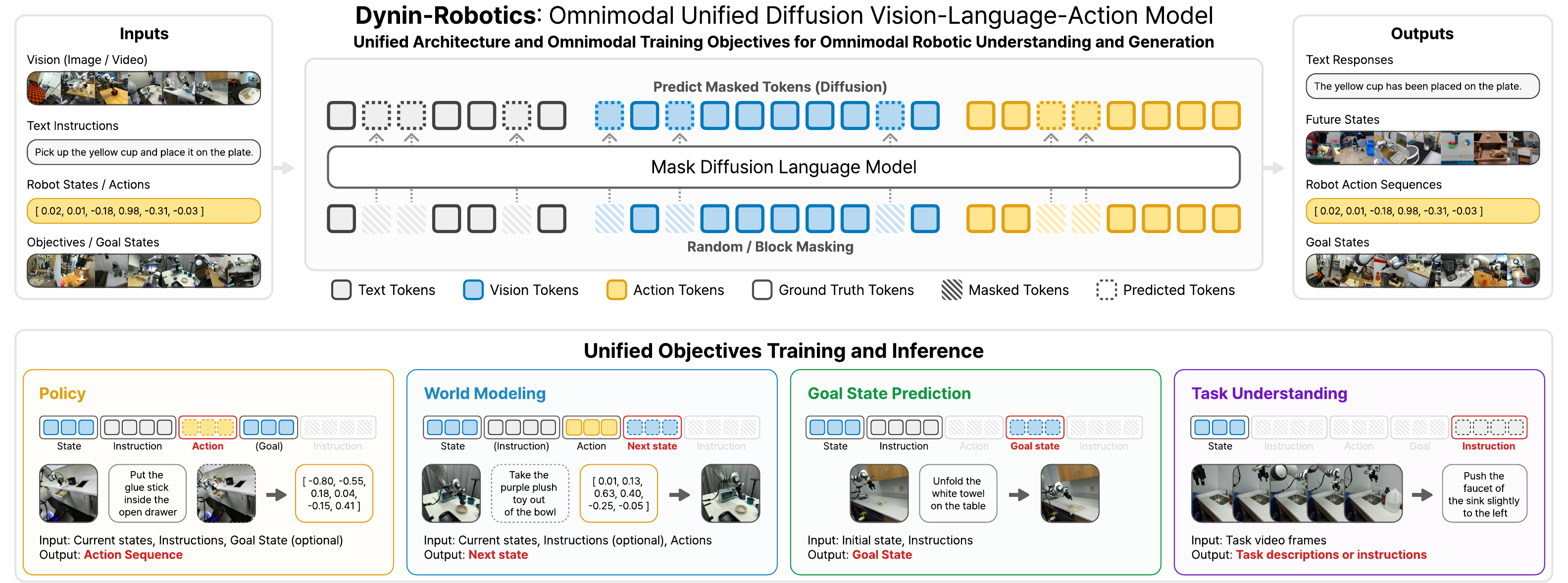}
\vspace{-1.0em}
% \caption{Overview of Dynin-Robotics.}
\caption{Overview of Dynin-Robotics. A single omnimodal masked-diffusion backbone represents language, image/video, and robot-action tokens in a shared sequence and reconstructs masked target spans. By changing the visible context, objective token, and masked target, the same model supports Policy, World Modeling, Goal-State Prediction, and Task Understanding, generating action chunks, next visual states, goal states, or task instructions.}
\label{fig:overview}
\end{figure}

\section{Introduction}
\label{sec:intro}

Vision-language-action (VLA) models~\cite{
brohan2023rt2,
kim2024openvla,
black2024pi0,
intelligence2025pi05,
bjorck2025gr00t,
zheng2025xvla}
connect natural-language instructions and visual observations to robot actions. Language-conditioned manipulation spans target identification, object interaction, and continuous control under contact, clutter, and occlusion~\cite{
baker2009action,
wolpert1995internal,
finn2017deep,
suomalainen2022survey,
poria202610,
kawaharazuka2025vision}. For example, following the instruction ``hand me something to cut the package'' involves selecting an object by its function and grasping it in the current scene. Visual goal and dynamics prediction offer representations of the intended outcome and the scene changes associated with an action. Bringing these predictions into action inference provides a way to connect task semantics with physical execution.

Recent work approaches this connection from different pretraining and modeling choices. VLM-based policies build on large vision-language backbones~\cite{
brohan2023rt2,
kim2024openvla,
black2024pi0,
intelligence2025pi05,
qu2025spatialvla,
zhao2025cot,
pertsch2025fast,
bjorck2025gr00t,
liang2025discretediffusionvla,
kim2025openvlaoft,
zheng2025xvla,
yang2026abot,
li2025cogvla}, adapting language--vision representations for instruction following and action prediction. Video-based policies and related predictive approaches~\cite{
sun2026vlajepa,
chen2025goalvla,
kim2026cosmos,
zhu2025uwm,
pai2025mimic,
liang2025videopolicy,
li2026lingbotva,
yuan2026fastwam}
incorporate future visual observations or latent scene representations into policy learning. These approaches emphasize different sources of supervision for linking language, visual changes, and control.

We examine how these modeling choices relate to policy behavior through a two-task diagnostic on VLABench~\cite{zhang2025vlabench}, comparing $\pi_{0.5}$~\cite{intelligence2025pi05} and Mimic-Video~\cite{pai2025mimic}. On SelectFruit, which involves target selection across object and layout variation, $\pi_{0.5}$ performs better across the evaluated instruction tracks. On InsertFlower, which involves grasping, alignment, and insertion, Mimic-Video is competitive particularly under the indirect Track~4 instructions. Replacing the instruction with a length-matched random string also produces a larger success-rate decrease for $\pi_{0.5}$ than for Mimic-Video. The two policies thus exhibit different task-performance profiles and different sensitivity to linguistic input. These observations motivate studying language conditioning and visual prediction together within a shared policy model.

Recent unified action models~\cite{
wang2025unifiedvla,
cen2025worldvla,
chen2025udvla,
wen2025dvla,
cen2025rynnvla002,
liu2026mmada}
combine action prediction with visual generation or trajectory understanding. Existing approaches offer several ways to connect predictions to control: VLM features can condition a dedicated action expert~\cite{black2024pi0,kim2025openvlaoft}, while unified prediction-and-understanding models combine visual information with action learning~\cite{zhang2025upvla}. We study a trajectory formulation in which instructions, observations, goals, and actions are represented as variables that can serve as either context or prediction targets. This formulation allows the same model to support several conditional learning tasks and to reuse their predictions through multiple inference compositions.

We introduce \method{}, a unified multimodal VLA foundation model built on Dynin-Omni~\cite{kim2026dynin}, an omnimodal masked-diffusion model. Dynin-Omni performs understanding and generation through iterative token prediction with a shared discrete interface and a single bidirectional backbone. \method{} extends this interface with quantized robot-action tokens and a reserved typed interface for aligned continuous sensor streams. As illustrated in Figure~\ref{fig:overview}, language instructions, visual observations, goal states, and robot actions occupy typed spans within a single trajectory sequence, with optional spans for sensor context.

Objective-conditioned masking specifies which trajectory spans remain visible and which the shared Transformer and prediction head reconstruct. \textsc{Policy} predicts action chunks from observations and instructions; \textsc{World Modeling} predicts the next visual observation conditioned on the current context and action; \textsc{Task Understanding} reconstructs an instruction from trajectory frames; and \textsc{Goal-State Prediction} predicts a terminal visual state corresponding to the instructed goal. Training these conditional queries within one representation connects actions with task language, local visual transitions, and terminal outcomes.

The learned interfaces also support inference composition. As summarized in Figure~\ref{fig:inference}, \method{} performs action-only decoding, predicts a goal before policy decoding, jointly denoises action and future-state spans, or uses \textsc{World Modeling} to score and rerank action candidates. Goal guidance can be combined with joint denoising or candidate filtering. These modes reuse the same model for policy, goal, and world-model queries. They support test-time scaling by allocating additional inference computation to visual prediction and candidate evaluation. Our experiments examine which compositions improve action-only performance and the throughput costs of those choices.

Block-parallel masked decoding reduces the sequential work required to generate an action chunk. At each denoising step, \method{} predicts masked action positions in parallel and progressively commits tokens. Combined with dInfer~\cite{ma2025dinferefficientinferenceframework}, an optimized inference stack for parallel decoding and context reuse, this improves model-side action-decoding throughput by up to $29.2\times$ under the reported profiling setup.

Our evaluation connects the training and inference design to downstream performance. On two VLABench tasks, robot pretraining improves adaptation within the same Stage-2 step budget, and the full objective mixture improves shifted-instruction success over Policy-only post-training under a fixed coupled decoder. With the checkpoint held fixed, selected inference compositions improve success over action-only decoding. We also evaluate the Stage-1 checkpoint on DROID~\cite{khazatsky2024droid} for action-conditioned visual prediction, goal-state generation, action prediction, and qualitative trajectory-to-instruction generation, without additional DROID-specific post-training. On policy benchmarks, \method{} achieves a 98.1\% average success rate on LIBERO~\cite{liu2023libero}, 73.0\% on zero-shot LIBERO-Plus~\cite{fei2025liberoplus}, and a 78.4\% average across four manipulation conditions on a Franka Research~3 platform.

Our main contributions are:

\begin{itemize}[left=0pt]

\item \textbf{Unified robot trajectory modeling with masked diffusion.}
We extend an omnimodal masked-diffusion backbone with robot-action tokens and formulate \textsc{Policy}, \textsc{World Modeling}, \textsc{Task Understanding}, and \textsc{Goal-State Prediction} as conditional denoising queries over a shared trajectory representation.

\item \textbf{Compositional and accelerated inference from a single model.}
The shared interface supports test-time scaling through goal prediction and action-candidate evaluation, with predicted visual states participating in action generation and selection. Block-parallel decoding and context reuse reduce sequential action-generation cost.

\item \textbf{Diagnostics and controlled evaluation of unified VLA modeling.}
A two-task VLABench diagnostic identifies differences in task performance and instruction sensitivity between $\pi_{0.5}$ and Mimic-Video. Within-model ablations evaluate robot pretraining, objective supervision, and inference composition, complemented by policy and multimodal-output evaluations in simulation and on a physical robot.

\end{itemize}

\section{Related Work}
\label{sec:rw}

\subsection{Large Diffusion Language Models}
\label{subsec:ldlm}

Most contemporary large language models~\cite{achiam2023gpt,grattafiori2024llama3,yang2024qwen2,dsv3} generate text autoregressively through left-to-right next-token prediction. Standard autoregressive decoding commits tokens in that order, with each prediction conditioned on the preceding sequence. Studies such as the Reversal Curse~\cite{berglund2024reversal} have also documented order-sensitive generalization behavior in these models. For robot trajectories, flexible generation order offers an alternative way to condition and predict related visual and action variables.

Diffusion models learn to reverse a data-corruption process~\cite{sohl2015deep,ho2020denoising}. Extensions to categorical variables use discrete transition processes~\cite{hoogeboom2021argmax,austin2021structured}, allowing this formulation to operate over token vocabularies.

Early diffusion language models applied continuous diffusion to token embeddings for controllable or sequence-to-sequence generation~\cite{li_diffusion-lm_2022,gong_diffuseq_2022}. Later methods model diffusion directly over discrete vocabularies. DiffusionBERT~\cite{he_diffusionbert_2023} uses an absorbing-state process based on token masking, while SEDD~\cite{pmlr-v235-lou24a} learns the ratios needed to reverse a discrete diffusion process.

Masked diffusion uses \texttt{[MASK]} as an absorbing noise state: the forward process replaces clean tokens with masks, and the learned reverse process reconstructs them from visible context. Each denoising step predicts the currently masked positions in parallel. Depending on the decoding schedule, uncertain predictions can be remasked and revised, allowing iterative refinement under bidirectional context.

MDLM~\cite{sahoo2024simple} develops a masked-diffusion language-modeling objective, and large-scale models including LLaDA~\cite{nie2025large} and Dream~\cite{ye2025dream} apply masked diffusion to language understanding, reasoning, and instruction following. These models provide a token-generation interface that supports parallel prediction and flexible refinement schedules.

Quantizing visual observations and continuous controls extends this interface to robot trajectories. An action chunk can be predicted and refined using context across its dimensions and temporal positions. Recent policies apply masked diffusion to action decoding~\cite{liang2025discretediffusionvla,wen2025llada} or jointly model actions with visual states and multimodal instructions~\cite{chen2025udvla,liu2026mmada}. Our formulation uses the shared discrete sequence to express robot control, visual prediction, and trajectory understanding as conditional denoising queries.

\subsection{Unified Multimodal Models}
\label{subsec:unified_multimodal_models}

Multimodal understanding and generation can share a language backbone while using different output interfaces. DreamLLM~\cite{dong2023dreamllm} and SEED-X~\cite{ge2024seed} connect multimodal language models with visual generators, while NeXT-GPT~\cite{next-gpt}, CoDi-2~\cite{CoDi-2}, and HyperCLOVAX-Omni~\cite{team2026hyperclova} extend generation to additional visual and speech modalities. These designs combine shared language processing with modality-specific generation components.

Another line of work represents multiple modalities in a shared token space
and models them using a common generative backbone. AnyGPT~\cite{anygpt},
Chameleon~\cite{team2024chameleon}, Emu3~\cite{wang2024emu3}, and
Janus-Pro~\cite{chen2025januspro} extend autoregressive language modeling to
multimodal token sequences. Beyond autoregressive modeling, BAGEL~\cite{bagle}
and Show-o2~\cite{xie2025showo2} combine language reasoning with diffusion- or
flow-based visual generation. In parallel, UniDisc~\cite{swerdlow2025unifiedmultimodaldiscretediffusion},
MMaDA~\cite{MMaDA}, Fudoki~\cite{wang2025fudoki},
LaViDa-O~\cite{li2025lavidao}, and Lumina-DiMOO~\cite{xin2025luminadimooomnidiffusionlarge}
explore discrete diffusion~\cite{sahoo2024simple,nie2025large} over shared multimodal tokens. More recently,
LLaDA2.0-Uni~\cite{ai2026llada20uniunifyingmultimodalunderstanding} combines
a discrete diffusion language-model backbone with a diffusion decoder to
support multimodal understanding, generation, and editing.

The shared-token approach provides a common representation for multimodal conditioning and prediction. Masked-diffusion models use visible tokens to condition target spans and refine those targets iteratively. Dynin-Omni~\cite{kim2026dynin} applies this approach to text, vision, and speech within a shared masked-denoising process. Dynin-Robotics extends the interface to robot trajectories, treating instructions, visual states, goals, and actions as context or targets for complementary prediction tasks.

\subsection{Modeling Paradigms in Robot Learning}
\label{subsec:modeling_paradigms_robot_learning}

\begin{figure}[t]
\centering
% \vspace{-0.3em}
\includegraphics[width=1\linewidth]{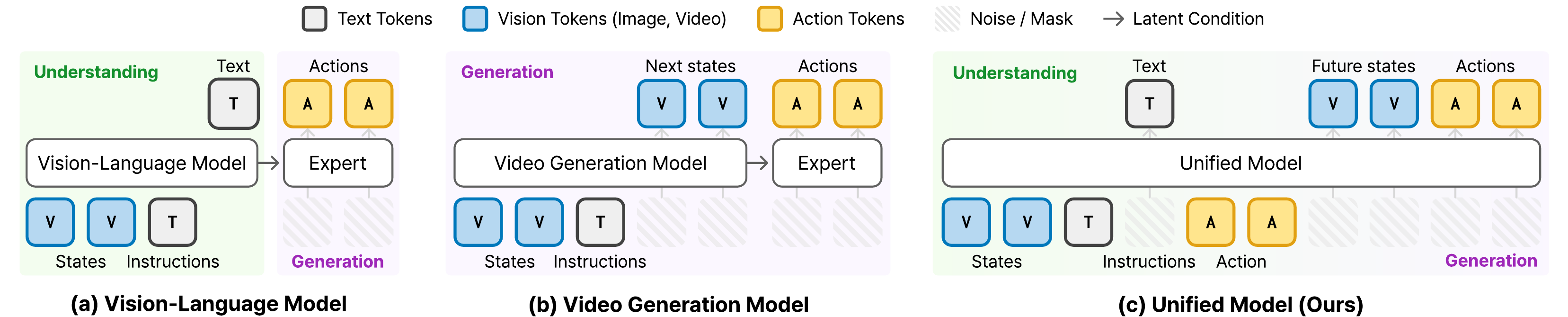}
\vspace{-1.0em}
% \caption{Robot-learning modeling paradigms. (a) Vision-language policies map visual states and instructions to actions through a dedicated action expert. (b) Video-generation policies predict future visual states and rely on a separate expert for action generation. (c) Dynin-Robotics uses a single masked-diffusion model to treat language, next visual states, and actions as either visible context or generation targets. Hatched spans denote masked or noisy targets, and arrows denote latent conditioning.}
\caption{Representative architectures for connecting perception, prediction, and robot actions. (a) A VLM conditions a dedicated action expert, a common design in recent VLM-based VLAs; other VLM-based policies generate actions directly. (b) A video generator conditions an action expert, illustrating one architecture for video-based control. (c) Dynin-Robotics represents language, visual states, and actions within a shared masked-diffusion model, with objective-specific choices of visible context and prediction targets. Hatched spans denote masked or noisy targets, and arrows denote latent conditioning.}
\label{fig:vla_architectures}
\end{figure}

\definecolor{adaptedcell}{RGB}{255,246,210}

\begin{table*}[t]
\centering
% \caption{
% Modeling paradigms of robot policies. \textit{NSP} and \textit{MoT} denote
% next-scale prediction and Mixture-of-Transformers. \textit{Modeling} indicates
% the dominant backbone formulation. The remaining columns compare video input,
% future prediction, world modeling, task understanding, future-guided control,
% and candidate reranking. Capabilities are marked only when explicitly
% supported. \(\ast\) denotes configuration-specific video input.
% }
% \caption{Capability comparison across robot-policy paradigms. \textit{NSP} and \textit{MoT} denote next-scale prediction and Mixture-of-Transformers; \textit{Modeling} gives the dominant backbone. Checkmarks and crosses indicate capabilities reported or not reported for the cited configuration, and \(\ast\) denotes configuration-dependent video input. The Task Understanding column follows source-reported capability and is broader than Dynin-Robotics' trajectory-to-instruction objective.}
\caption{Reported capabilities of robot-policy models. Modeling summarizes the architectural or generative formulation associated with each cited configuration. NSP and MoT denote next-scale prediction and Mixture-of-Transformers, respectively. Checkmarks and crosses indicate capabilities reported or not reported for the cited configuration, and an asterisk denotes configuration-dependent video input. Task Understanding follows the capability described in each source and encompasses a broader range of language-understanding tasks than Dynin-Robotics' trajectory-to-instruction objective.}
\label{tab:unified-comparison}

\scriptsize
\setlength{\tabcolsep}{5pt}
\renewcommand{\arraystretch}{0.91}

\begin{tabular}{lccccccc}
\toprule
Model &
Modeling &
\shortstack{Video\\input} &
\shortstack{Goal-state\\prediction} &
\shortstack{World\\model} &
\shortstack{Task\\understanding} &
\shortstack{Goal-state\\guided policy} &
\shortstack{Candidate\\reranking} \\
\midrule

\multicolumn{8}{l}{\textit{VLM-Based Policies}} \\
RT-2~\cite{brohan2023rt2}
  & AR
  & \cmark\textsuperscript{*} & \xmark & \xmark
  & \cmark & \xmark & \xmark \\

OpenVLA~\cite{kim2024openvla}
  & AR
  & \xmark & \xmark & \xmark & \xmark & \xmark & \xmark \\

OpenVLA-OFT~\cite{kim2025openvlaoft}
  & AR
  & \cmark\textsuperscript{*} & \xmark & \xmark
  & \xmark & \xmark & \xmark \\

SpatialVLA~\cite{qu2025spatialvla}
  & AR
  & \xmark & \xmark & \xmark & \xmark & \xmark & \xmark \\

CoT-VLA~\cite{zhao2025cot}
  & AR
  & \xmark & \cmark & \xmark & \xmark & \cmark & \xmark \\

CogVLA~\cite{li2025cogvla}
  & AR
  & \xmark & \xmark & \xmark & \xmark & \xmark & \xmark \\

$\pi_0$~\cite{black2024pi0}
  & Flow
  & \xmark & \xmark & \xmark & \xmark & \xmark & \xmark \\

$\pi_0$-FAST~\cite{pertsch2025fast}
  & AR
  & \xmark & \xmark & \xmark & \xmark & \xmark & \xmark \\

$\pi_{0.5}$~\cite{intelligence2025pi05}
  & Flow
  & \xmark & \xmark & \xmark & \cmark & \xmark & \xmark \\

GR00T-N1, N1.6~\cite{bjorck2025gr00t}
  & Flow
  & \xmark & \xmark & \xmark & \xmark & \xmark & \xmark \\

F1~\cite{lv2025f1}
  & MoT + NSP
  & \xmark & \cmark & \xmark & \xmark & \cmark & \xmark \\

InternVLA-M1%
~\cite{internvlam1}
  & AR
  & \xmark & \xmark & \xmark & \xmark & \xmark & \xmark \\

Discrete Diffusion VLA~\cite{liang2025discretediffusionvla}
  & Masked Diff.
  & \xmark & \xmark & \xmark & \xmark & \xmark & \xmark \\

LLaDA-VLA~\cite{wen2025llada}
  & Masked Diff.
  & \xmark & \xmark & \xmark & \xmark & \xmark & \xmark \\

X-VLA~\cite{zheng2025xvla}
  & Flow
  & \xmark & \xmark & \xmark & \xmark & \xmark & \xmark \\

ABot-M0~\cite{yang2026abot}
  & DiT
  & \xmark & \xmark & \xmark & \xmark & \xmark & \xmark \\

NORA~\cite{hung2025nora}
  & AR
  & \xmark & \xmark & \xmark & \xmark & \xmark & \xmark \\

RIPT-VLA~\cite{tan2025ript}
  & AR
  & \xmark & \xmark & \xmark & \xmark & \xmark & \xmark \\

\midrule
\multicolumn{8}{l}{\textit{World-/Video-Model-Based Policies}} \\
Video Policy~\cite{liang2025videopolicy}
  & Cont.\ Diff.
  & \xmark & \cmark & \xmark & \xmark & \cmark & \xmark \\

Cosmos Policy~\cite{kim2026cosmos}
  & Cont.\ Diff.
  & \xmark & \cmark & \cmark & \xmark & \xmark & \cmark \\

Mimic-Video~\cite{pai2025mimic}
  & Flow
  & \xmark & \cmark & \xmark & \xmark & \cmark & \xmark \\

LingBot-VA~\cite{li2026lingbotva}
  & AR Diff.
  & \cmark & \cmark & \cmark & \xmark & \cmark & \xmark \\

Fast-WAM~\cite{yuan2026fastwam}
  & Cont.\ Diff.
  & \xmark & \cmark & \xmark & \xmark & \xmark & \xmark \\

WorldVLA~\cite{cen2025worldvla}
  & AR
  & \xmark & \cmark & \cmark & \xmark & \xmark & \xmark \\

RynnVLA-002~\cite{cen2025rynnvla002}
  & AR
  & \cmark & \cmark & \cmark & \xmark & \xmark & \xmark \\

UWM~\cite{zhu2025uwm}
  & Cont.\ Diff.
  & \xmark & \cmark & \cmark & \xmark & \xmark & \xmark \\

\midrule
\multicolumn{8}{l}{\textit{Unified Policies}} \\
UniVLA~\cite{wang2025unifiedvla}
  & AR
  & \cmark & \cmark & \xmark & \xmark & \xmark & \xmark \\

UP-VLA~\cite{zhang2025upvla}
  & AR
  & \xmark & \cmark & \xmark & \xmark & \xmark & \xmark \\

UD-VLA~\cite{chen2025udvla}
  & Masked Diff.
  & \xmark & \cmark & \xmark & \xmark & \cmark & \xmark \\

MMaDA-VLA~\cite{liu2026mmada}
  & Masked Diff.
  & \xmark & \cmark & \xmark & \xmark & \cmark & \xmark \\

dVLA~\cite{wen2025dvla}
  & Masked Diff.
  & \xmark & \xmark & \xmark & \xmark & \xmark & \xmark \\

\rowcolor{lightblue}
\method{} (Ours)
  & Masked Diff.
  & \cmark & \cmark & \cmark & \cmark & \cmark & \cmark \\

\bottomrule
\end{tabular}
\end{table*}

Robot-learning approaches differ in how they use language--vision representations, visual prediction, and action generation. Figure~\ref{fig:vla_architectures} illustrates representative VLM-based, video-based, and unified architectures. The categories describe modeling emphasis, with several methods combining elements of more than one. Table~\ref{tab:unified-comparison} summarizes the cited configurations in terms of video context, visual prediction, task understanding, and inference-time action guidance.

\paragraph{VLM-based policies.}
Vision-language-model-based (\emph{VLM-based}) policies adapt language--vision representations to robot control. RT-2~\cite{brohan2023rt2} and OpenVLA~\cite{kim2024openvla} represent actions as language-like tokens, whereas $\pi_0$~\cite{black2024pi0} and $\pi_{0.5}$~\cite{intelligence2025pi05} use flow-matching action experts for continuous control. The expert-based architecture in Figure~\ref{fig:vla_architectures}(a) is widely used in recent VLM-based VLAs; direct action-token generation provides another output interface. Other works improve spatial grounding, intermediate reasoning, adaptation efficiency, or preservation of pretrained VLM capabilities~\cite{qu2025spatialvla,zhao2025cot,li2025cogvla,kim2025openvlaoft}. Discrete Diffusion VLA~\cite{liang2025discretediffusionvla} and LLaDA-VLA~\cite{wen2025llada} apply discrete diffusion to action decoding, introducing parallel token refinement within VLM-based policies.

\paragraph{VGM-based policies.}
Video-generative-model-based (\emph{VGM-based}) policies and related world-action models use visual prediction in policy learning. Cosmos Policy~\cite{kim2026cosmos} and Mimic-Video~\cite{pai2025mimic} adapt pretrained video models for control, while LingBot-VA~\cite{li2026lingbotva} jointly models frame prediction and policy execution. Fast-WAM~\cite{yuan2026fastwam} examines the roles of test-time future imagination and video co-training. The video-generator-plus-expert design in Figure~\ref{fig:vla_architectures}(b) illustrates one way to connect visual generation to actions; the cited methods use different mechanisms to couple the two outputs.

\paragraph{Coupled and unified policies.}
Unified approaches combine action learning with visual prediction or understanding. WorldVLA~\cite{cen2025worldvla}, RynnVLA-002~\cite{cen2025rynnvla002}, and UWM~\cite{zhu2025uwm} integrate action prediction and world modeling, linking the world-/video-model-based and unified approaches summarized in Table~\ref{tab:unified-comparison}. UD-VLA~\cite{chen2025udvla} and MMaDA-VLA~\cite{liu2026mmada} jointly denoise visual states and actions. UP-VLA~\cite{zhang2025upvla} combines multimodal understanding, future visual prediction, and action learning, while dVLA~\cite{wen2025dvla} incorporates multimodal chain-of-thought prediction into action modeling.

These approaches connect visual prediction to action learning through different conditioning structures. In our formulation, action-conditioned \textsc{World Modeling} predicts a next observation given an action, while \textsc{Goal-State Prediction} predicts a terminal visual target from an observation and instruction. \method{} combines these queries with \textsc{Policy} and trajectory-to-instruction \textsc{Task Understanding} within one masked-diffusion backbone. Their shared interface supports goal conditioning, joint action--next-state denoising, and action-candidate evaluation through changes to the visible context and prediction targets.
\section{Dynin-Robotics}
\label{sec:dynin_robotics}

Dynin-Robotics formulates robot learning as conditional masked denoising over shared trajectories containing visual observations, language instructions, robot actions, and optional next-state images, goal images, sensor streams, and metadata. An objective token specifies the visible context and target spans. Policy, World Modeling, Task Understanding, and Goal-State Prediction use different conditioning configurations of this representation. The same interface supports the six inference compositions in Section~\ref{subsec:inference}, including the default two-stage path that predicts a goal-state context before denoising an action chunk.

\subsection{Backbone}
\label{subsec:backbone}

Dynin-Robotics builds on Dynin-Omni~\cite{kim2026dynin}, an omnimodal masked-diffusion model. Dynin-Omni represents text, images, sampled video frames, and speech as discrete tokens and processes their interleaved sequences with a single bidirectional Transformer, shared token embeddings, and a unified categorical prediction head. This architecture supports multimodal understanding and generation through the same backbone.

Each inherited modality uses a pretrained tokenizer: a language tokenizer for text, a shared MAGViT-v2 tokenizer~\cite{yu2023magvit} for images and sampled video frames, and the EMOVA-Speech-Tokenizer~\cite{chen2024emova} for speech. The shared visual vocabulary represents both observed and generated images. The inherited text and visual tokenizers and their corresponding detokenizers remain frozen during robot training.

Dynin-Omni is trained to reconstruct masked tokens from visible multimodal context~\cite{sahoo2024simple,nie2025large,kim2026dynin}. Changing the visible and target spans allows a modality to condition another prediction or become a generation target itself. Bidirectional attention conditions masked positions on the visible sequence, and iterative denoising reconstructs the targets. For robot trajectories, this interface supports predictions in several conditional directions, including actions from observations and instructions, visual outcomes from actions, and instructions from trajectory frames.

\begin{figure}[t]
\centering
\vspace{-0.3em}
\includegraphics[width=1\linewidth]{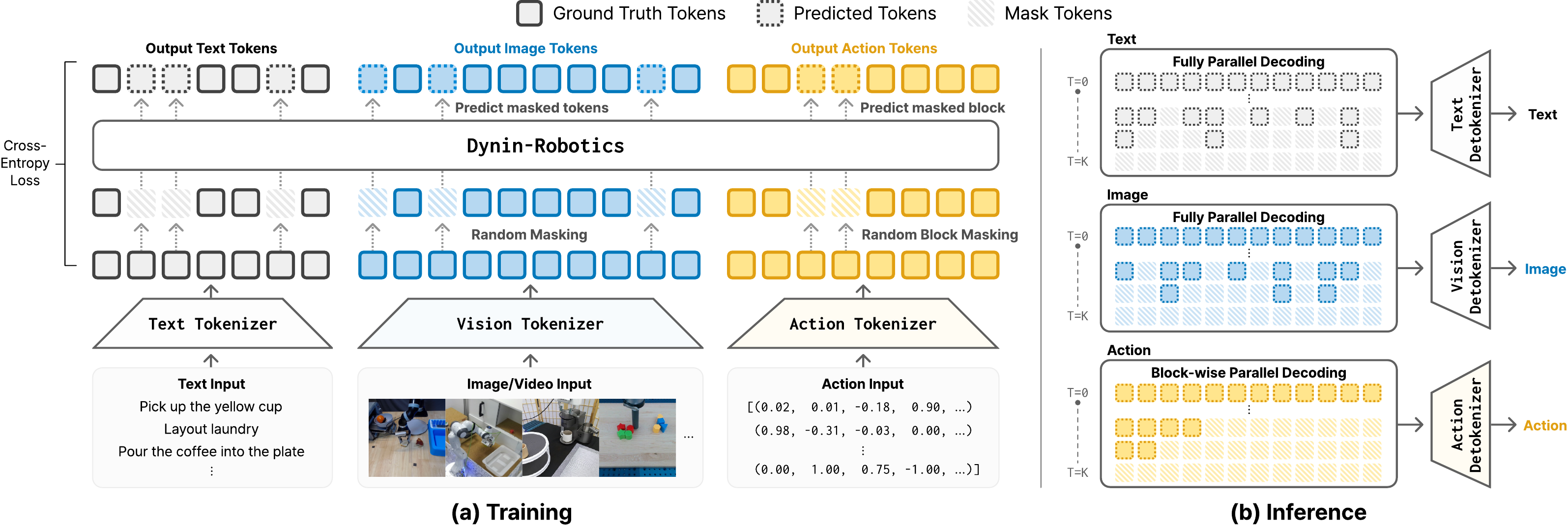}
\vspace{-1.0em}
% \caption{Unified masked-diffusion backbone and modality decoding. (a) During training, text, image/video, and continuous actions are converted into discrete token sequences. The shared backbone reconstructs randomly masked text and visual tokens and block-masked action tokens using a common cross-entropy objective. (b) At inference, text and visual targets are decoded fully in parallel, whereas action chunks are reconstructed through block-wise parallel denoising and mapped back to continuous controls by the action detokenizer.}
\caption{Shared masked-diffusion backbone and modality-specific token interfaces. (a) Text and visual observations are tokenized, and continuous actions are discretized through uniform quantization. For each training objective, the backbone reconstructs masked positions within the designated target spans while retaining the conditioning context. Text and visual targets use random masking, and action targets use block masking. (b) Inference iteratively refines masked targets, with parallel prediction at each denoising step and block-wise commitment for actions. Inverse quantization maps the resulting action tokens to continuous controls.}
\label{fig:architecture_dynin_robotics}
\end{figure}

As illustrated in Figure~\ref{fig:architecture_dynin_robotics}, we retain the Dynin-Omni Transformer and text--visual tokenization interface and extend its discrete token space to robot actions. Speech tokens remain part of the inherited checkpoint, but speech-conditioned objectives are inactive in the reported robotics runs. The following sections define the robot token space and objective-specific conditioning layouts.

\subsection{Robotic Token Space Extension}
\label{subsec:robot_token_space}

For notation, let \(\mathcal{V}_{\mathrm{text}}^{+}\) include lexical text tokens together with the objective, delimiter, padding, termination, and \texttt{[MASK]} control tokens allocated in the text-tokenizer range, and let \(M_S\) denote the number of sensor-type token blocks reserved by the configuration. The integrated Dynin-Robotics token space is
\begin{equation}
\begin{aligned}
\mathcal{V}_{\mathrm{DR}}
&= \mathcal{V}_{\mathrm{text}}^{+}
\uplus \mathcal{V}_{\mathrm{vision}}
\uplus \mathcal{V}_{\mathrm{speech}} \uplus \mathcal{V}_{\mathrm{action}}
\uplus \mathcal{V}_{\mathrm{sensor}}, \qquad
\mathcal{V}_{\mathrm{sensor}}
=\biguplus_{m=1}^{M_S}\mathcal{V}_{S_m}.
\end{aligned}
\end{equation}
where \(\uplus\) denotes non-overlapping token-ID blocks. The inherited speech block is present in the checkpoint but inactive in the reported robotics runs. The sets \(\mathcal{V}_{\mathrm{vision}}\) and \(\mathcal{V}_{\mathrm{action}}\) contain image/video-frame and robot-action tokens, respectively; when continuous sensor streams are configured, \(\mathcal{V}_{S_m}\) denotes a reserved block for sensor type \(m\).

For objective \(o\), a robot example follows the typed serialization template
\begin{equation}
\mathbf{x}^{(o)}=
[o;\mathbf{x}^{V};\mathbf{x}^{T};\mathbf{x}^{A};\mathbf{x}^{V}_{t+1};\mathbf{x}^{V}_{g};\mathbf{x}^{S};\mathbf{x}^{M}],
\end{equation}
where \(\mathbf{x}^{V}\) denotes observation or trajectory-frame tokens, \(\mathbf{x}^{T}\) denotes instruction or textual target tokens, \(\mathbf{x}^{A}\) denotes action tokens, \(\mathbf{x}^{V}_{t+1}\) denotes optional immediately subsequent visual-state tokens, \(\mathbf{x}^{V}_{g}\) denotes optional terminal or successful goal-state tokens, \(\mathbf{x}^{S}\) denotes one or more optional typed sensor spans, and \(\mathbf{x}^{M}\) denotes optional textual metadata encoded and delimited within \(\mathcal{V}_{\mathrm{text}}^{+}\). The active spans, visible context, targets, and masking pattern are selected for each objective. Next-state and goal-state spans are included when used by that objective or inference composition, as illustrated in the objective-layout figure below.

\subsubsection{Actions}
\label{subsec:actions}

For an action chunk \(\mathbf{a}_{t:t+H-1}\in\mathbb{R}^{H\times d_a}\), the reported Dynin-Robotics model uses fixed, deterministic, per-dimension uniform binning; no learned action tokenizer is used. Source-specific preprocessing first maps native controls to the common action interface described in Section~\ref{subsec:training_data}. For channel \(j\), robust lower and upper statistics \(\ell^{A}_{j}\) and \(u^{A}_{j}\) are computed from the corresponding training data. The Stage-1 OXE preprocessing uses the per-source 1st and 99th percentiles for the six motion channels. Let \(\epsilon>0\) be a numerical floor for degenerate normalization ranges, and let \(v^{A}_{\mathrm{start}}\) denote the first global token ID of the reserved action-token block. The action is normalized, quantized into \(B_A\) active bins, and mapped into the reserved action-token block as
\begin{align}
\bar a_{t+h,j}
&=\operatorname{clip}\!\left(
2\frac{a_{t+h,j}-\ell^{A}_{j}}
{\max(u^{A}_{j}-\ell^{A}_{j},\epsilon)}-1,
-1,1\right), \nonumber\\
b^{A}_{t+h,j}
&=\operatorname{clip}\!\left(
\left\lfloor\frac{B_A(\bar a_{t+h,j}+1)}{2}\right\rfloor,
0,B_A-1\right), \nonumber\\
x^{A}_{t+h,j}
&=v^{A}_{\mathrm{start}}+b^{A}_{t+h,j}.
\label{eq:action_binning}
\end{align}
Here, \(h\in\{0,\ldots,H-1\}\) and \(j\in\{1,\ldots,d_a\}\). The \(Hd_a\) action tokens are serialized in time-major order. The six motion channels in the reported 7-DoF interface use source- or benchmark-specific robust normalization, while the gripper is retained as an absolute channel rather than accumulated as a delta. Constant or unused motion dimensions are set to zero before tokenization.

After denoising, each predicted bin is mapped to its normalized bin center and then returned to continuous action space:
\[
\begin{aligned}
\widehat{\bar a}_{t+h,j}
&=2\frac{\widehat b^{A}_{t+h,j}+\tfrac12}{B_A}-1,\\
\widehat a_{t+h,j}
&=\ell^{A}_{j}
+\frac{\widehat{\bar a}_{t+h,j}+1}{2}
(u^{A}_{j}-\ell^{A}_{j}).
\end{aligned}
\]
The released Stage-1 checkpoint uses \(B_A=256\), whereas Stage 2 selects a task-specific active bin count within the same reserved action block; Section~\ref{sec:sensitivity} evaluates this choice. All reported experiments use uniform binning.

\subsubsection{Optional Sensor Token Interface}
\label{subsec:sensors}

The same deterministic discretization interface can extend to aligned continuous robot-sensor streams. For sensor type \(m\) with a window \(\mathbf{s}^{(m)}\in\mathbb{R}^{H_m\times d_m}\), preprocessing synchronizes or resamples the stream to the observation timestamps before tokenization. Every channel is then calibrated with configuration-specific robust lower and upper statistics \(\ell^{(m)}_j\) and \(u^{(m)}_j\), computed from the corresponding training split; the optional force/torque preprocessing path uses channel-wise 1st and 99th percentiles:
\begin{equation}
\bar s^{(m)}_{t+h,j}
=\operatorname{clip}\!\left(
2\frac{s^{(m)}_{t+h,j}-\ell^{(m)}_j}
{\max(u^{(m)}_j-\ell^{(m)}_j,\epsilon)}-1,
-1,1\right).
\label{eq:sensor_normalization}
\end{equation}
Here, \(h\in\{0,\ldots,H_m-1\}\). The normalized value is uniformly quantized into \(B_{S_m}\) bins and mapped to the non-overlapping reserved token block \(\mathcal{V}_{S_m}\), whose first global token ID is denoted by \(v^{S_m}_{\mathrm{start}}\):
\begin{align}
b^{(m)}_{t+h,j}
&=\operatorname{clip}\!\left(
\left\lfloor\frac{B_{S_m}(\bar s^{(m)}_{t+h,j}+1)}{2}\right\rfloor,
0,B_{S_m}-1\right), \nonumber\\
x^{S_m}_{t+h,j}
&=v^{S_m}_{\mathrm{start}}+b^{(m)}_{t+h,j}.
\label{eq:sensor_binning}
\end{align}
Within each typed sensor span, tokens are serialized in time-major order over the \(H_m d_m\) values. Sensor-specific channel statistics absorb differences in unit, range, and scale, while \(H_m\), \(d_m\), the typed delimiters, and the fixed channel order encode temporal extent, dimensionality, and sensor identity. Multiple streams can be represented by concatenating non-overlapping typed visible spans without changing the masked-diffusion objective; when a stream is unavailable, its optional span is omitted. This interface can accommodate calibrated vector-valued signals such as force/torque, proprioception, and tactile measurements.

The sensor interface reserves capacity for optional sensor conditioning. Sensor-token training is inactive in the released 220k Stage-1 checkpoint, and the reported Stage-2 experiments use no sensor-conditioned policy decoding. Sensor detokenization and sensor-generation objectives are also inactive in the reported setup.

\subsection{Unified Objective Training}
\label{subsec:training}

\begin{figure}[t]
\centering
\vspace{-0.3em}
\includegraphics[width=1\linewidth]{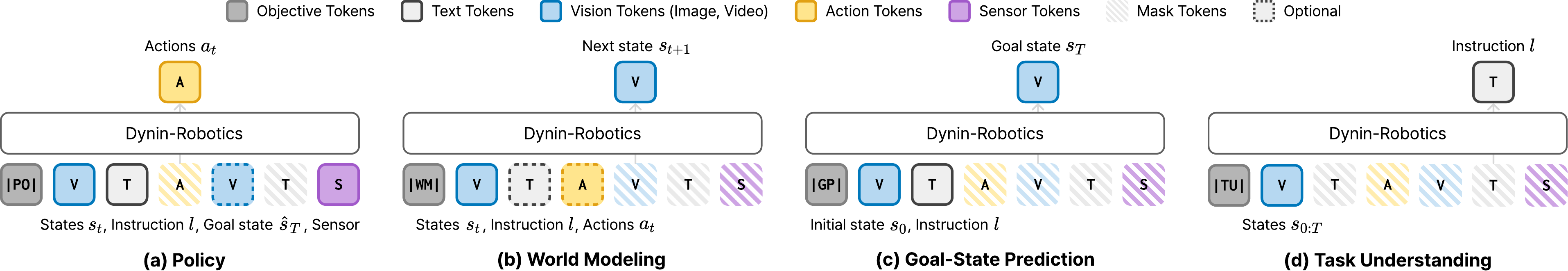}
\vspace{-1.0em}
\caption{Objective-specific observed–target layouts used in unified training. (a) Policy predicts action tokens from states, an instruction, and optional goal-state and sensor context. (b) World Modeling predicts next-state visual tokens conditioned on states, the instruction, and actions. (c) Goal-State Prediction predicts a terminal goal state from the initial state and instruction. (d) Task Understanding reconstructs the task instruction from trajectory states. Solid tokens are visible context, hatched tokens are masked prediction targets, and dotted spans are optional.}
\label{fig:unified_objective_training}
\end{figure}

Dynin-Robotics constructs objective-specific examples from a common corpus of robot trajectories. Let \(\mathcal{O}=\{o_{\mathrm{PO}},o_{\mathrm{WM}},o_{\mathrm{TU}},o_{\mathrm{GP}}\}\) denote Policy, World Modeling, Task Understanding, and Goal-State Prediction. For objective \(o\), let \(\mathcal{C}_o\) be the visible conditioning positions and \(\mathcal{I}_o\) the target positions in \(\mathbf{x}^{(o)}\). The model learns the conditional target distribution
\[
p_{\theta}\!\left(\mathbf{x}^{(o)}_{\mathcal{I}_o}\mid
\mathbf{x}^{(o)}_{\mathcal{C}_o},o\right)
\]
using the objective-specific prompt layouts in Figure~\ref{fig:unified_objective_training}.

The Policy objective predicts an action chunk from the current visual state and instruction, with goal-state and sensor context included when those spans are available. Goal-State Prediction predicts a terminal visual state corresponding to the instructed goal from the initial observation and instruction. World Modeling predicts the immediately subsequent visual state \(s_{t+1}\) conditioned on the current visual context \(s_t\), instruction, and corresponding action chunk. Throughout this work, World Modeling denotes this action-conditioned next-state visual-token prediction task. Task Understanding reconstructs the instruction from sampled trajectory frames. The objectives thus associate action chunks with local visual transitions, terminal outcomes, and task descriptions within the same trajectory representation.

For each sampled objective, corruption is applied only inside the selected target span or spans, while conditioning tokens remain visible. Text and visual targets use random token masking, whereas action targets use contiguous block masking over the time-major action sequence. For an objective-\(o\) micro-batch, let \(N_o\) denote the number of examples. For example \(b\), let \(\mathcal{I}_{o,b}\) denote its target positions and \(\widetilde{\mathbf{x}}^{(o)}_b\) the corrupted input sequence after target-only masking. The sampled mask probability \(\rho_b\) masks at least one target token in example \(b\); unmasked target tokens are copied into the input and excluded from supervision. Let \(L_{o,b}=|\mathcal{I}_{o,b}|\) be the full target-span length and
\[
\mathcal{M}_{o,b}
=\{i\in\mathcal{I}_{o,b}:\widetilde x^{(o)}_{b,i}=\texttt{[MASK]}\}
\]
be the masked target positions. Before cross-entropy is computed, the unified prediction head is sliced to the vocabulary of the target modality. The target-only masked-diffusion loss is
\begin{equation}
\begin{split}
\mathcal{L}_{\mathrm{MDM}}(o)
=\mathbb{E}\!\Bigg[
&-\frac{1}{N_o}\sum_{b=1}^{N_o}
\frac{1}{\rho_b L_{o,b}}\times\sum_{i\in\mathcal{M}_{o,b}}
\log p^{(v_o)}_{\theta}\!\left(
x^{(o)}_{b,i}\mid\widetilde{\mathbf{x}}^{(o)}_b,o
\right)
\Bigg],
\end{split}
\label{eq:mdm_loss}
\end{equation}
where \(v_o\) denotes the target-modality vocabulary. Inverse mask-probability weighting makes the estimator comparable across sampled mask rates, while normalization by the full target length prevents examples with more masked positions from receiving systematically larger weight. For the Stage-1 Policy objective, let \(N_{\mathrm{PO}}\) denote the number of Policy examples in the micro-batch, and define the normalized center of action bin \(r\) as \(\gamma_r=2(r+\tfrac12)/B_A-1\). The expected normalized action and Policy-only auxiliary are
\begin{align}
\widehat{\bar a}_{b,i}
&=\sum_{r=0}^{B_A-1}
p^{(A)}_{\theta,i}\!\left(
r\mid\widetilde{\mathbf{x}}^{(\mathrm{PO})}_b,o_{\mathrm{PO}}
\right)\gamma_r, \nonumber\\
\mathcal{L}_{\mathrm{bin}}
&=\frac{1}{N_{\mathrm{PO}}}
\sum_{b=1}^{N_{\mathrm{PO}}}
\frac{1}{|\mathcal{M}^{A}_b|}
\sum_{i\in\mathcal{M}^{A}_b}
\left|\widehat{\bar a}_{b,i}-\bar a_{b,i}\right|.
\end{align}
where \(\mathcal{M}^{A}_b\) contains the masked, unpadded action positions and \(\bar a_{b,i}\) is the corresponding flattened normalized action target. Section~\ref{subsec:continual_pretraining} specifies the Stage-1 coefficient applied to this Policy-only auxiliary.

At training micro-step \(n\), let \(\mathcal{A}_n\) denote the scheduled objective multiset and let \(\mathcal{L}^{(n)}_{\mathrm{MDM}}(o)\) denote the corresponding micro-batch estimate. The common objective-mixture loss is
\begin{equation}
\mathcal{L}^{(n)}_{\mathrm{mix}}
=\frac{1}{|\mathcal{A}_n|}
\sum_{o\in\mathcal{A}_n}
\lambda_o\mathcal{L}^{(n)}_{\mathrm{MDM}}(o).
\label{eq:unified_loss}
\end{equation}
Here, repeated entries in \(\mathcal{A}_n\) determine objective exposure, whereas \(\lambda_o\) controls loss scaling. The Stage-1 and Stage-2 objectives in Section~\ref{sec:training_recipe} specialize this expression with their respective schedules, weights, and auxiliary losses.

Training proceeds in two stages. Stage 1 performs unified continual pretraining on large-scale multimodal robot trajectories. Stage 2 post-trains the same token interface and masked-denoising objective on benchmark-specific robot data. Dataset mixtures, action-tokenization settings, objective schedules, and optimization details are reported in Section~\ref{sec:training_recipe}.

\subsection{Unified Multi-Stage Inference}
\label{subsec:inference}

\begin{figure}[t]
\centering
\vspace{-0.3em}
\includegraphics[width=1.0\linewidth]{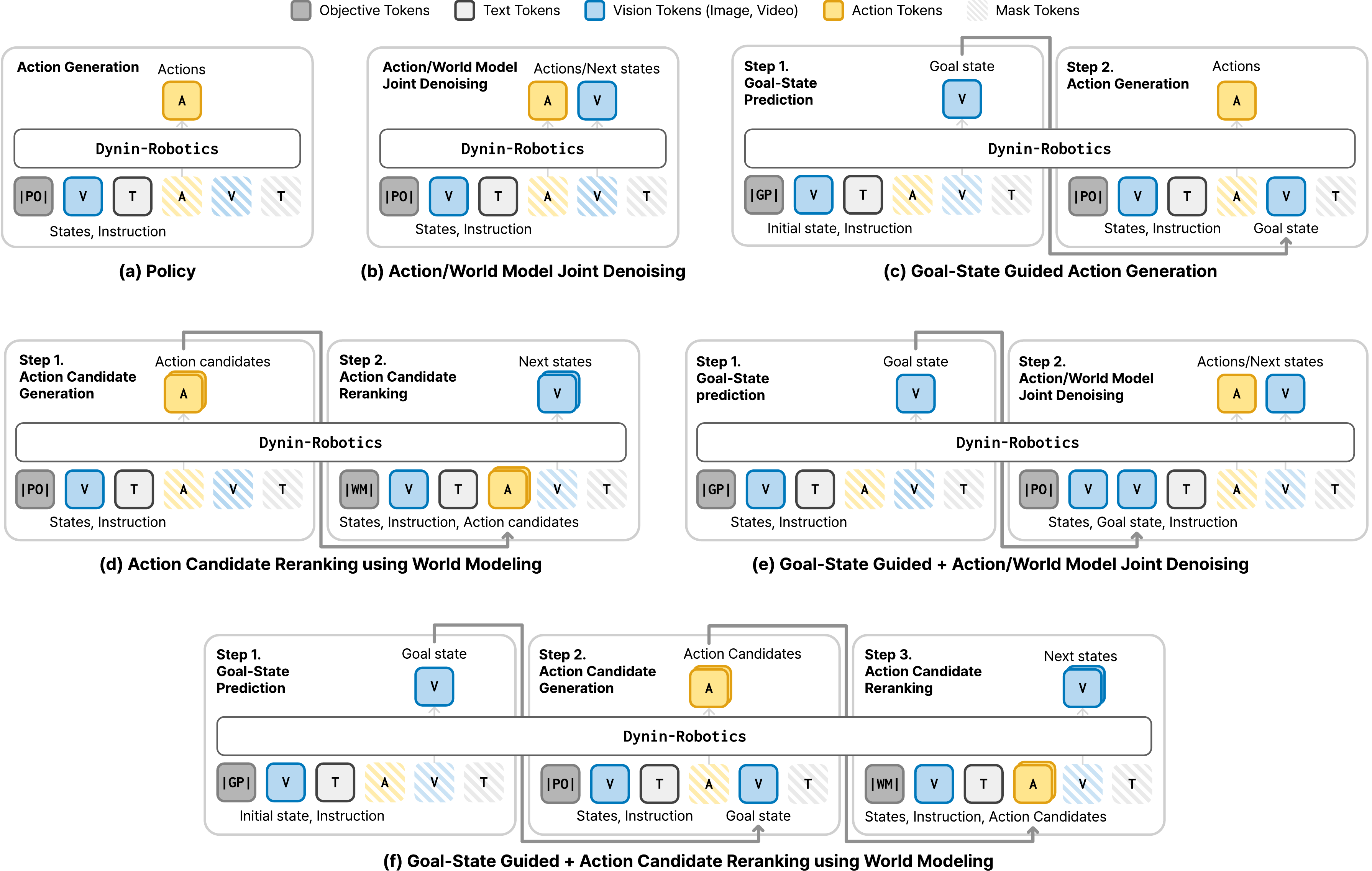}
\vspace{-0.3em}
\caption{Six inference compositions from the shared denoising interface: (a) action-only Policy; (b) joint action/next-state denoising; (c) the default two-stage goal-guided Policy; (d) world-model action-candidate reranking; (e) goal-guided joint denoising; and (f) goal-guided candidate reranking. Hatched spans are denoising targets, and solid spans are visible context.}
\label{fig:inference}
\end{figure}

Figure~\ref{fig:inference} summarizes the six inference compositions. For individual objective queries, Dynin-Robotics uses the same objective-conditioned prompts at inference as in training; the joint modes combine trained target spans through an explicitly identified inference-time composite mask. Given visible context and an objective token, each target span is initialized with \texttt{[MASK]} tokens and decoded by reverse masked diffusion. We use \(k\in\{K,\ldots,0\}\) exclusively for the denoising index: \(k=K\) denotes a fully masked target and \(k=0\) a fully decoded target. We use \(\ell\) to index the language instruction, so \(\mathbf{x}^{T}_{\ell}\) denotes its text-token span. At each step, the model predicts all currently masked target positions, commits high-confidence tokens, and refines the remaining low-confidence positions. Visual targets are decoded in parallel, whereas action targets are refined in contiguous blocks over the time-major action sequence.

\paragraph{Action-only and goal-state-guided action decoding.}
The action-only policy mode denoises an action chunk from the current observation and instruction:
\[
\widehat{\mathbf{x}}^{A,(0)}_{t:t+H-1}
\sim p_{\theta}\!\left(
\mathbf{x}^{A}_{t:t+H-1}\mid
\mathbf{x}^{V}_{t},\mathbf{x}^{T}_{\ell},o_{\mathrm{PO}}
\right).
\]
When goal-state guidance is enabled, Dynin-Robotics first predicts a goal-state context from the initial observation and language instruction:
\begin{equation}
\widehat{\mathbf{x}}^{V}_{g}
\sim p_{\theta}\!\left(
\mathbf{x}^{V}_{g}\mid
\mathbf{x}^{V}_{0},\mathbf{x}^{T}_{\ell},o_{\mathrm{GP}}
\right),
\label{eq:goal_prediction}
\end{equation}
and then provides the predicted goal-state tokens as visible context for Policy decoding:
\begin{equation}
\widehat{\mathbf{x}}^{A,(0)}_{t:t+H-1}
\sim p_{\theta}\!\left(
\mathbf{x}^{A}_{t:t+H-1}\mid
\mathbf{x}^{V}_{t},\mathbf{x}^{T}_{\ell},
\widehat{\mathbf{x}}^{V}_{g},o_{\mathrm{PO}}
\right).
\label{eq:goal_guided_policy}
\end{equation}
In either mode, the generated action tokens are mapped from their predicted bins to continuous controls using the inverse quantization defined in Section~\ref{subsec:actions}, and are executed with receding-horizon replanning. The goal-state-guided two-stage path is the default inference configuration, whereas action-only decoding is retained as an ablation.

\paragraph{Joint action--world-model denoising.}
The shared token interface also supports an inference-time composite mask in which the action span and action-conditioned next-state visual span are initialized together and iteratively refined under the same visible context. Let \(\mathcal{I}_{\mathrm{PO}}\equiv\mathcal{I}_{o_{\mathrm{PO}}}\) and \(\mathcal{I}_{\mathrm{WM}}\equiv\mathcal{I}_{o_{\mathrm{WM}}}\) denote the Policy and World Modeling target positions, respectively, within this composite layout. With
\(
\mathcal{I}_{\mathrm{J}}=\mathcal{I}_{\mathrm{PO}}\cup\mathcal{I}_{\mathrm{WM}}
\), let \(\widetilde{\mathbf{x}}_{\mathrm{J}}^{(K)}\) denote the inference-only composite sequence containing the shared visible context, the trained Policy and World Modeling objective markers, and fully masked action and next-state target spans. The joint decoding is
\begin{equation}
\left(
\widehat{\mathbf{x}}^{A,(0)}_{t:t+H-1},
\widehat{\mathbf{x}}^{V,(0)}_{t+1}
\right)
=\operatorname{Decode}_{\theta}\!\left(
\widetilde{\mathbf{x}}_{\mathrm{J}}^{(K)};
\mathcal{I}_{\mathrm{PO}},
\mathcal{I}_{\mathrm{WM}}
\right).
\label{eq:joint_inference}
\end{equation}
The optional goal-state span \(\widehat{\mathbf{x}}^{V}_{g}\) is included in the visible context of \(\widetilde{\mathbf{x}}_{\mathrm{J}}^{(K)}\) only in the goal-guided variant. This inference-only composite reuses the trained objective markers, prediction head, and model parameters within one query. At each denoising step, the shared head is restricted to the action vocabulary on \(\mathcal{I}_{\mathrm{PO}}\) and to the visual vocabulary on \(\mathcal{I}_{\mathrm{WM}}\).

\paragraph{World-model-based candidate reranking.}
Dynin-Robotics uses the World Modeling objective to evaluate candidate actions at inference. Starting from the action-only or goal-guided Policy output \(\widehat{\mathbf{x}}^{A,(0)}\), we form a local candidate set \(\mathcal{C}\) by replacing low-confidence action tokens with high-probability alternatives under the original Policy logits. The shared model scores these candidates, and an acceptance rule determines whether to replace the original output.

Before scoring, we construct one reference visual-token span \(\mathbf{y}\) and hold it fixed across all candidates. In the goal-guided filtering mode, \(\mathbf{y}=\widehat{\mathbf{x}}^{V}_{g}\). In a next-state filtering mode, \(\mathbf{y}\) is predicted once from the base Policy output \(\mathbf{c}^{0}\) and then reused as the reference for every candidate. The World Modeling energy is the mean negative log-likelihood assigned to this common reference:
\begin{equation}
E_{\mathrm{WM}}(\mathbf{c})
=-\frac{1}{|\mathcal{Y}|}
\sum_{j\in\mathcal{Y}}
\log p_{\theta}\!\left(
y_j\mid
\mathbf{x}^{V}_{t},\mathbf{x}^{T}_{\ell},
\mathbf{c},o_{\mathrm{WM}}
\right),
\label{eq:wm_energy}
\end{equation}
where \(\mathcal{Y}\) is the common set of reference visual-target positions. Every candidate is scored against the same reference \(\mathbf{y}\).

For action position \(i\), let \(k_i\) denote the denoising step at which the candidate alternatives for that position are collected, and let \(\widetilde{\mathbf{x}}^{(\mathrm{PO},k_i)}\) denote the corresponding partially denoised Policy input. We define the frozen base-Policy distribution
\begin{equation}
\pi^{\mathrm{base}}_i(v)
=p^{(A)}_{\theta,i}\!\left(
v\mid
\widetilde{\mathbf{x}}^{(\mathrm{PO},k_i)},
\mathbf{x}^{V}_{t},
\mathbf{x}^{T}_{\ell},
[\widehat{\mathbf{x}}^{V}_{g}],
o_{\mathrm{PO}}
\right).
\label{eq:base_policy_distribution}
\end{equation}
The distributions \(\pi^{\mathrm{base}}_i\) are computed once and reused for all candidates. A Policy prior penalizes candidates with low probability under these distributions:
\begin{equation}
\begin{aligned}
E_{\mathrm{PO}}(\mathbf{c})
&=-\frac{1}{L_A}\sum_{i=1}^{L_A}
\log \pi^{\mathrm{base}}_i(c_i), \\
J_{\mathrm{rank}}(\mathbf{c})
&=E_{\mathrm{WM}}(\mathbf{c})
+\alpha_{\mathrm{prior}}E_{\mathrm{PO}}(\mathbf{c})
\end{aligned}
\label{eq:rerank_score}
\end{equation}
where \(L_A=Hd_a\), \(\alpha_{\mathrm{prior}}\geq0\) is the Policy-prior weight, and the bracketed goal-state context is optional. We select
\(
\mathbf{c}^{*}=\arg\min_{\mathbf{c}\in\mathcal{C}}J_{\mathrm{rank}}(\mathbf{c})
\), but accept it only when it improves both the world-model energy by a margin \(\delta\) and the combined ranking score:
\begin{equation}
\begin{aligned}
E_{\mathrm{WM}}(\mathbf{c}^{0})-E_{\mathrm{WM}}(\mathbf{c}^{*})
&\geq\delta, \\
J_{\mathrm{rank}}(\mathbf{c}^{0})-J_{\mathrm{rank}}(\mathbf{c}^{*})
&>0
\end{aligned}
\label{eq:rerank_gate}
\end{equation}
where \(\mathbf{c}^{0}=\widehat{\mathbf{x}}^{A,(0)}\) is the original Policy output. If either condition is not satisfied, \(\mathbf{c}^{0}\) is executed. The World Modeling query therefore supplies an acceptance criterion for local alternatives to the Policy output. We evaluate this branch as an ablation.

These compositions support test-time scaling through additional prediction and evaluation queries to the shared model. Goal guidance allocates computation to a visual target before action decoding, joint denoising refines action and next-state spans together, and filtering allocates computation to candidate evaluation. Combining these operations provides several ways to trade inference computation for policy performance.

\subsection{Acceleration}
\label{subsec:method_acceleration}

\begin{figure}[t]
\centering
\vspace{-0.3em}
\includegraphics[width=1\linewidth]{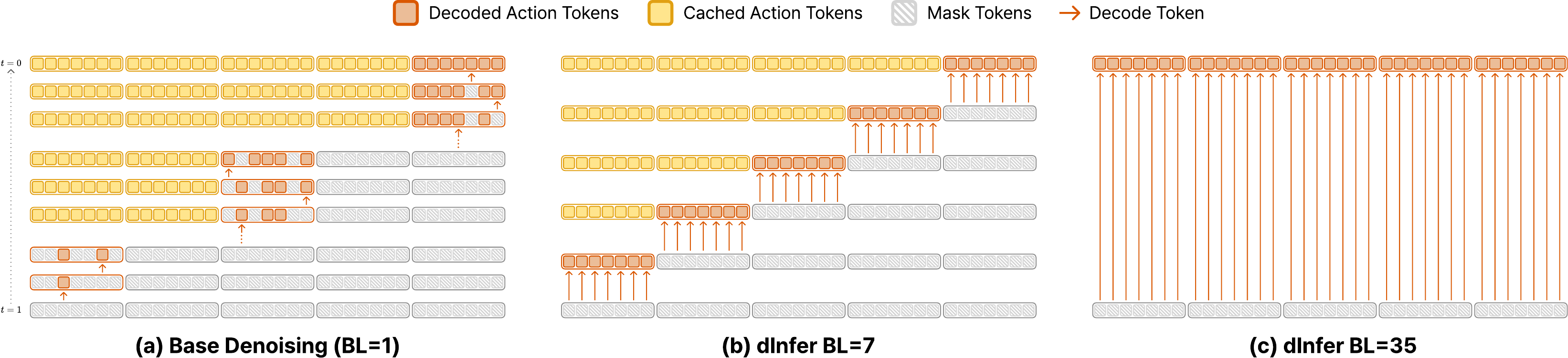}
\vspace{-1.0em}
% \caption{Block-parallel denoising for a five-step, 7-DoF action chunk. The chunk is serialized as 35 time-major tokens; a block length of 7 covers one action timestep, whereas a block length of 35 covers the full chunk. Orange, yellow, and hatched boxes denote newly decoded, cached, and masked tokens.}
\caption{Schematic token-commitment patterns for a five-step, 7-DoF action chunk serialized as 35 time-major tokens. Block lengths of 1, 7, and 35 correspond to individual tokens, one action timestep, and the full chunk, respectively. Orange tokens are newly committed, yellow tokens are retained from earlier steps, and hatched tokens remain masked. Confidence-based commitment determines which positions require further refinement. The schematics illustrate block granularity; realized backbone evaluations are reported in Table~\ref{tab:dinfer_control_tradeoff}.}
\label{fig:block_size}
\end{figure}

Dynin-Robotics predicts all masked action positions in parallel during each backbone evaluation. Confidence-based commitment determines which predictions become part of the decoded sequence and which positions receive further refinement. Figure~\ref{fig:block_size} illustrates a five-step, 7-DoF chunk serialized as 35 time-major tokens. A block length of 7 spans one action timestep, while a block length of 35 spans the full chunk. The configured block length and the realized number of backbone evaluations are distinct quantities; Section~\ref{sec:decoding_sensitivity} evaluates their latency and control trade-offs.

The accelerated implementation reuses computation for the fixed visual--language--goal context across denoising steps, following diffusion-LLM caching strategies~\cite{wu2026fastdllm,ma2025dinferefficientinferenceframework}. This reuse is approximate because changes to target tokens affect the bidirectional backbone. Candidate scoring can also be batched when actions share the observation, instruction, target-span length, and decoding schedule. The base and dInfer~\cite{ma2025dinferefficientinferenceframework} decoders use confidence-based commitment schedules. The online reinforcement-learning stage in Section~\ref{subsubsec:rlft} replaces that schedule with a learned count-and-position remasking controller while keeping the shared token predictor and modality tokenizers frozen. These implementation choices target the latency of model-side action generation.

\section{Training Recipe}
\label{sec:training_recipe}

\subsection{Training Data}
\label{subsec:training_data}

\begin{figure*}[t]
    \centering
    \includegraphics[width=\textwidth]{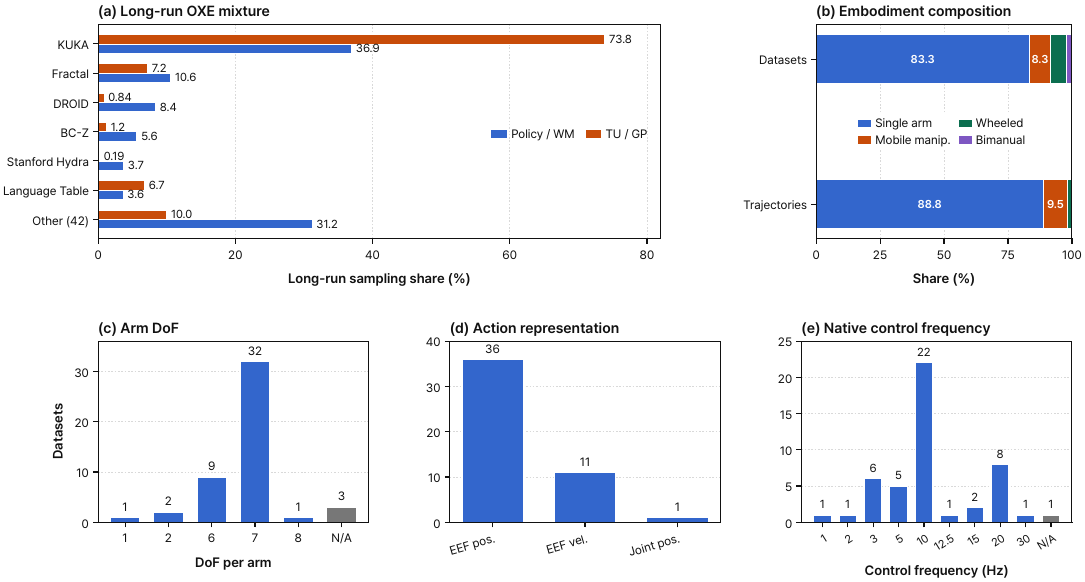}
    \caption{Composition of the 48-dataset Stage-1 Open X-Embodiment mixture. (a) Long-run sampling mass averaged across partitions and virtual epochs; Policy and World Modeling are weighted by transitions, whereas Task Understanding and Goal-State Prediction are weighted by trajectories. (b) Embodiment shares under dataset- and trajectory-count denominators. (c--e) Dataset counts by arm DoF, native action representation, and control frequency. Panels use different denominators; all sources map to the common 7D action interface.}
    \label{fig:oxe_stage1_profile}
\end{figure*}

\paragraph{Continual-pretraining data.}
Our Stage-1 mixture contains 48 Open X-Embodiment (OXE) datasets~\cite{openxembodiment2023}, comprising 1,332,985 trajectories and 65,217,081 transitions in the runtime training split. The collection spans single-arm, mobile-manipulator, wheeled, and bimanual source embodiments, with heterogeneous kinematics and control rates (Figure~\ref{fig:oxe_stage1_profile}). Before mixing, source-specific transforms map native controls to a common 7D end-effector interface containing 3D translation, 3D rotation, and gripper control. The six motion channels are normalized with per-source robust quantiles, the gripper remains an absolute channel, and constant or unused motion dimensions are set to zero before action tokenization. The active bimanual source is reduced to the same 7D action view. Source embodiment DoF describes the diversity of the collected data; the model's action output uses the common 7D interface.

The optional force/torque preprocessing path implements the sensor interface described in Section~\ref{subsec:sensors}. The released 220k Stage-1 run uses only the OXE mixture above, with no active force/torque or force-policy batches; its corpus statistics and evaluation therefore cover that mixture.

\paragraph{Post-training data.}
Stage~2 uses separate domain-specific runs for the four standard LIBERO suites~\cite{liu2023libero}, the SelectFruit and InsertFlower tasks of VLABench~\cite{zhang2025vlabench}, and the real-world Franka manipulation tasks described in Section~\ref{subsec:eval_setup}. Each run has its own training data and objective schedule. Domain-specific adapters change camera mappings, trajectory indexing, action statistics, and optional textual metadata while preserving the shared objective-conditioned token interface. DROID results use the Stage-1 checkpoint on the DROID validation set without additional DROID-specific post-training.

\subsection{Continual Pretraining}
\label{subsec:continual_pretraining}

Starting from the Dynin-Omni initialization, we continually pretrain the shared masked-diffusion Transformer on robot sequences while keeping the inherited text and visual tokenizers frozen. Following Section~\ref{subsec:training}, let
\(
\mathcal{O}=\{o_{\mathrm{PO}},o_{\mathrm{WM}},o_{\mathrm{TU}},o_{\mathrm{GP}}\}
\)
denote Policy, World Modeling, Task Understanding, and Goal-State Prediction. Every objective uses the target-only masked-diffusion loss \(\mathcal{L}_{\mathrm{MDM}}(o)\) defined there. We reserve \(k\) for the denoising index and use \(n\) for the training micro-step. Let \(\mathcal{L}^{(n)}_{\mathrm{MDM}}(o)\) and \(\mathcal{L}^{(n)}_{\mathrm{bin}}\) denote the corresponding micro-batch estimates at training micro-step \(n\). A deterministic schedule forms an objective multiset \(\mathcal{A}_n\), which may contain repeated Policy requests. The Stage-1 micro-step loss is
\begin{equation}
\begin{aligned}
\mathcal{L}^{(n)}_{\mathrm{Stage\text{-}1}}
&=\frac{1}{|\mathcal{A}_n|}
\sum_{o\in\mathcal{A}_n}\lambda_o
\Bigl[
\mathcal{L}^{(n)}_{\mathrm{MDM}}(o)
+\mathbb{I}[o=o_{\mathrm{PO}}]\cdot\beta\mathcal{L}^{(n)}_{\mathrm{bin}}
\Bigr].
\end{aligned}
\label{eq:stage1_loss}
\end{equation}
Here, \(\mathbb{I}[\cdot]\) is the indicator function and \(\beta=0.2\). Schedule multiplicity determines long-run objective exposure, whereas \(\lambda_o\) controls loss scaling within a micro-step. These two quantities are varied independently. The expected-bin MAE auxiliary \(\mathcal{L}^{(n)}_{\mathrm{bin}}\) is applied only to the Stage-1 Policy objective.

\paragraph{Balanced distributed rotation.}
For objective \(o\), source \(i\) has effective sampling mass \(\mu_i^{(o)}=w_iN_i^{(o)}\), where \(w_i\) is its mixture weight and \(N_i^{(o)}\) counts transitions for Policy or World Modeling and trajectories for Task Understanding or Goal-State Prediction. Large sources are first divided into virtual partitions, producing 99 partition records from the 48 base datasets. The records are balanced into eight source groups. In the reported 32-rank run, ranks are assigned modulo eight to eight data lanes, with four ranks per lane. At virtual epoch \(e\), each lane visits the next group in a lane-specific cyclic order; ranks within the lane consume disjoint, epoch-permuted record shards before trajectory decoding and augmentation. Thus, every lane covers all eight balanced groups over eight virtual epochs while avoiding redundant host-side preprocessing.

Training proceeds in two consecutive segments. The first segment performs 21.8k optimizer updates with equal per-objective loss weights. The second segment loads only the model weights from that checkpoint, reinitializes the optimizer and scheduler, and performs a further 5.8k updates with a more policy-centered schedule and
\[
(\lambda_{\mathrm{PO}},\lambda_{\mathrm{WM}},\lambda_{\mathrm{TU}},\lambda_{\mathrm{GP}})
=(1,0.2,0.05,0.2).
\]
Both segments use AdamW with a peak learning rate of \(2\times10^{-5}\), \((\beta_1,\beta_2)=(0.9,0.95)\), weight decay \(0.01\), cosine decay, 4k warm-up updates, gradient accumulation of eight, gradient clipping at \(1.0\), and BF16 training with TF32 enabled. The two segments total 27.6k optimizer updates, or 220.8k accumulated micro-steps; the released checkpoint name reflects this micro-step count.

\subsection{Post-Training}
\label{subsec:post_training}

Stage 2 adapts the continual-pretraining checkpoint separately to each downstream domain \(d\). It preserves the shared token vocabulary, objective-conditioned prompt layouts, and masked-denoising interface, while recomputing action-normalization statistics and selecting an active action-bin count for the target domain. Objective schedules and loss weights are checkpoint-specific: a run may use Policy-only supervision or a unified four-objective multiset when the corresponding targets are available. Similarly, \(B_A\) is selected per downstream setting within the reserved action block. Section~\ref{sec:sensitivity} evaluates \(B_A\in\{16,32,64\}\) for the VLABench Stage-2 study and selects \(B_A=32\) for that comparison; the released Stage-1 checkpoint instead uses \(B_A=256\).

\subsubsection{Trajectory Metadata Conditioning}
\label{subsubsec:metadata}

For Stage-2 Policy examples with metadata enabled, we augment the instruction with demonstration-side text fields describing a normalized local action-smoothness proxy and trajectory length. Let \(\{\mathbf{u}_h\}_{h=0}^{4}\), \(\mathbf{u}_h\in\mathbb{R}^{6}\), denote the first six normalized arm channels of a five-step action chunk; the discontinuous gripper channel is excluded. We reconstruct a normalized arm path as
\[
\begin{aligned}
\mathbf{y}_h&=\sum_{r=0}^{h}\mathbf{u}_r
&&\text{for delta actions},\\
\mathbf{y}_h&=\mathbf{u}_h
&&\text{for absolute actions}.
\end{aligned}
\]
For control interval \(\Delta t\), the two available third finite differences and their mean norm are
\begin{equation}
\begin{aligned}
\mathbf{j}_h
&=\frac{\mathbf{y}_h-3\mathbf{y}_{h-1}+3\mathbf{y}_{h-2}-\mathbf{y}_{h-3}}
{\Delta t^3},
\quad h\in\{3,4\}, 
\quad J_{\mathrm{chunk}}
=\frac{1}{2}\sum_{h=3}^{4}\|\mathbf{j}_h\|_2.
\end{aligned}
\label{eq:trajectory_jerk}
\end{equation}
We use \(\Delta t=0.05\) s for LIBERO and VLABench and approximately \(0.0667\) s for the 15-Hz real-world setting. Because the calculation is performed in normalized action coordinates and contains only two jerk vectors, \(J_{\mathrm{chunk}}\) is a local action-chunk proxy rather than physical Cartesian jerk in SI units.

Let \(\tau\) denote the complete successful demonstration trajectory and let \(L_{\tau}=|\tau|\) denote its length in environment steps. The visible text context serializes the instruction together with the fields \texttt{Action-Chunk Jerk} (\(J_{\mathrm{chunk}}\)), \texttt{Trajectory Steps} (\(L_{\tau}\)), and \texttt{Success} before the target action span. The success field is fixed to \texttt{true} for the successful demonstrations in the training source. These quantities use the existing text-token interface. At rollout time, the prompt instead supplies fixed values: unless overridden, \texttt{Action-Chunk Jerk} is set to 300 and \texttt{Trajectory Steps} to 200, together with the desired \texttt{Success} value. This conditioning uses no ground-truth future actions or realized episode lengths.

\subsubsection{Supervised Fine-Tuning}
\label{subsubsec:sft}

SFT reuses the target-only masked-diffusion loss defined in Section~\ref{subsec:training}. Let \(\mathcal{A}^{(d)}_n\) denote the active-objective multiset scheduled at Stage-2 micro-step \(n\) for domain \(d\), and let \(\lambda_{d,o}\geq0\) denote the domain- and objective-specific loss weight. For each scheduled request \(o\in\mathcal{A}^{(d)}_n\), its conditioning span remains visible, only a sampled subset of its target span is corrupted, and logits are restricted to the target-modality vocabulary. Let \(\mathcal{L}^{(n,d)}_{\mathrm{MDM}}(o)\) denote the corresponding micro-batch estimate of \(\mathcal{L}_{\mathrm{MDM}}(o)\). The Stage-2 micro-step loss is
\begin{equation}
\mathcal{L}^{(n,d)}_{\mathrm{SFT}}
=\frac{1}{|\mathcal{A}^{(d)}_n|}
\sum_{o\in\mathcal{A}^{(d)}_n}
\lambda_{d,o}\mathcal{L}^{(n,d)}_{\mathrm{MDM}}(o).
\label{eq:stage2_sft}
\end{equation}
For a Policy-only run, \(\mathcal{A}^{(d)}_n\) contains only \(o_{\mathrm{PO}}\); unified runs retain the same backbone and denoising interface while changing visible context, target modality, schedule exposure, and \(\lambda_{d,o}\). Stage 2 uses categorical masked-denoising without the Stage-1 expected-bin MAE auxiliary unless explicitly stated otherwise.

Let \(N_d\) denote the number of Stage-2 micro-steps for domain \(d\), let \(\mathcal{B}^{(d)}_n\) denote the sampled domain-specific micro-batch at step \(n\), and let \(\xi_n\) collect masking and corruption randomness. Initializing \(\theta\) from the Stage-1 model weights, we optimize
\begin{equation}
\theta_d^{\star}
=\arg\min_{\theta}
\frac{1}{N_d}
\sum_{n=1}^{N_d}
\mathbb{E}_{\mathcal{B}^{(d)}_n,\xi_n}
\left[
\mathcal{L}^{(n,d)}_{\mathrm{SFT}}
\left(\theta;\mathcal{B}^{(d)}_n,\xi_n\right)
\right].
\label{eq:stage2_objective}
\end{equation}
The objective multiset \(\mathcal{A}^{(d)}_n\) remains deterministically specified by the checkpoint-specific schedule.

\subsubsection{Online Reinforcement-Learning Fine-Tuning}
\label{subsubsec:rlft}

Supervised fine-tuning reconstructs demonstrated action tokens but does not directly optimize online task return or the number of backbone evaluations used during iterative decoding. We therefore introduce a separate online reinforcement-learning stage for the remasking schedule. Prior work has treated diffusion denoising as a sequential decision process~\cite{black2024trainingdiffusion,ren2025diffusionpolicy}, optimized diffusion initialization or step allocation~\cite{wagenmaker2025steering,yu2025d3p}, and applied reinforcement learning to discrete-diffusion VLAs~\cite{wu2026dvlarl}. To isolate schedule adaptation, the Dynin-Robotics backbone \(p_{\theta}\) and all modality tokenizers remain frozen; only a lightweight remasking controller \(\pi_{\phi}\) and its value estimator are trainable.

\paragraph{Remasking decision process.}
At environment step \(t\) and inner denoising step \(k\), the controller observes
\begin{equation}
\boldsymbol{\zeta}_{t,k}
=\left(
\mathbf{x}^{V}_{t},
\mathbf{x}^{T}_{\ell},
\mathbf{x}^{M}_{t},
\widehat{\mathbf{x}}^{V}_{g},
o_{\mathrm{PO}},
\mathbf{x}^{A,(k)}_{t:t+H-1},
k
\right),
\label{eq:online_rl_state}
\end{equation}
where the metadata span \(\mathbf{x}^{M}_{t}\) and goal-state span \(\widehat{\mathbf{x}}^{V}_{g}\) are omitted when their corresponding conditioning paths are disabled. For metadata-conditioned online rollouts, \(\mathbf{x}^{M}_{t}\) uses the fixed placeholder fields defined in Section~\ref{subsubsec:metadata} and remains visible throughout the inner denoising process. Decoding begins from a fully masked action span of length \(L_A=Hd_a\). Let
\[
\mathcal{M}_{t,k}
=\left\{
i\in\{1,\ldots,L_A\}:
x^{A,(k)}_{t,i}=\texttt{[MASK]}
\right\}
\]
denote the currently masked positions. The frozen backbone proposes tentative values \(\bar{x}^{A,(k-1)}_{t,i}\) for all \(i\in\mathcal{M}_{t,k}\) in parallel and exposes their hidden states and categorical distributions to the controller.

\paragraph{Count-and-position controller.}
Let \(\mathbf{h}_{\theta,t,k,i}\) denote the frozen backbone hidden state at action-token position \(i\), and let \(p_{\theta,t,k,i}(v)\) denote its categorical distribution over active action-token values \(v\in\mathcal{V}_{\mathrm{action}}\) at denoising step \(k\). For each tentative prediction, the controller constructs
\begin{equation}
\begin{aligned}
\mathbf{f}_{t,k,i}=\bigl[
&\mathbf{h}_{\theta,t,k,i};
\max_v p_{\theta,t,k,i}(v);
\mathsf{H}(p_{\theta,t,k,i});\\
&p^{(1)}_{\theta,t,k,i}-p^{(2)}_{\theta,t,k,i};
k/K
\bigr],
\end{aligned}
\label{eq:online_rl_features}
\end{equation}
where \(\mathsf{H}\) is categorical entropy and \(p^{(1)}\) and \(p^{(2)}\) are the two largest token probabilities. A position head assigns the score \(z_{t,k,i}=g^{\mathrm{pos}}_{\phi}(\mathbf{f}_{t,k,i})\). We collect these scores as
\(
\mathbf{z}_{t,k}=(z_{t,k,i})_{i\in\mathcal{M}_{t,k}}.
\)
To avoid an exponentially large subset action space, a count head first samples the number of positions to remask, \(c_{t,k}\in\{0,\ldots,|\mathcal{M}_{t,k}|-1\}\), and the position head then samples an ordered set \(\mathbf{r}_{t,k}\) of that size without replacement according to the scores. With controller action \(u_{t,k}=(c_{t,k},\mathbf{r}_{t,k})\), its likelihood factorizes as
\begin{equation}
\begin{aligned}
\log\pi_{\phi}(u_{t,k}\mid\boldsymbol{\zeta}_{t,k})
={}&\log\pi^{\mathrm{cnt}}_{\phi}
(c_{t,k}\mid\boldsymbol{\zeta}_{t,k})\\
&+\log\pi^{\mathrm{pos}}_{\phi}
(\mathbf{r}_{t,k}\mid c_{t,k},\mathbf{z}_{t,k}).
\end{aligned}
\label{eq:online_rl_policy}
\end{equation}
The strict upper bound on \(c_{t,k}\) guarantees that at least one new action token is accepted at every inner step; \(c_{t,k}=0\) accepts all tentative predictions and terminates denoising. For \(\mathcal{R}_{t,k}\) equal to the positions in \(\mathbf{r}_{t,k}\), the next token state is
\begin{equation}
x^{A,(k-1)}_{t,i}
=
\begin{cases}
\texttt{[MASK]}, & i\in\mathcal{R}_{t,k},\\
\bar{x}^{A,(k-1)}_{t,i}, & i\in\mathcal{M}_{t,k}\setminus\mathcal{R}_{t,k},\\
x^{A,(k)}_{t,i}, & i\notin\mathcal{M}_{t,k}.
\end{cases}
\label{eq:online_rl_transition}
\end{equation}
When the inner process terminates, inverse action quantization maps \(\mathbf{x}^{A,(0)}_{t:t+H-1}\) to continuous controls, which are executed with receding-horizon replanning.

\paragraph{Efficiency-aware online objective.}
The controller trades environment return against iterative-decoding cost. Let \(T_{\mathrm{env}}\) denote the environment rollout horizon, \(\gamma\in[0,1]\) the discount factor, and \(s_t\) the environment state at step \(t\). Let \(\widehat{\mathbf{a}}_t\) denote the control actually executed after decoding and inverse quantization, and let \(r_{\mathrm{env}}(s_t,\widehat{\mathbf{a}}_t,s_{t+1})\) denote the resulting environment reward. If \(K_t\leq K\) is the number of backbone evaluations used to decode the action chunk at environment step \(t\), the objective is
\begin{equation}
\begin{aligned}
J_{\mathrm{RL}}(\phi)
=\mathbb{E}_{\pi_{\phi}}\!\Biggl[
\sum_{t=0}^{T_{\mathrm{env}}-1}\gamma^t\Bigl(&
r_{\mathrm{env}}(s_t,\widehat{\mathbf{a}}_t,s_{t+1})-\eta_{\mathrm{den}}K_t-\eta_{\mathrm{step}}
\Bigr)\Biggr],
\end{aligned}
\label{eq:online_rl_objective}
\end{equation}
where the expectation covers controller actions and environment transitions, \(\eta_{\mathrm{den}}\) penalizes additional refinement, and \(\eta_{\mathrm{step}}\) discourages unnecessarily long episodes. We maximize \(J_{\mathrm{RL}}(\phi)\) with respect to the controller parameters \(\phi\). We optimize the controller and value estimator with proximal policy optimization (PPO)~\cite{schulman2017proximal} while keeping \(p_{\theta}\) frozen. At inference, the count head selects the most likely \(c_{t,k}\), and the \(c_{t,k}\) positions with the largest remasking scores are reconsidered. This learned controller replaces only the refinement schedule; it does not modify the shared token predictor or modality tokenizers. The present ablations do not isolate the controller's contribution from the other post-training and decoding choices.

\section{Evaluation}
\label{sec:eval}

\subsection{Evaluation Setup}
\label{subsec:eval_setup}

Our evaluation examines how the shared trajectory formulation performs as a robot policy, as a predictor of visual states and task descriptions, and as an interface for composing predictions during control. We first assess policy success on standard tasks, distribution shifts, and physical manipulation. We then evaluate the model's direct outputs and use ablations to distinguish the effects of robot pretraining, Stage-2 objective mixtures, and inference composition. Finally, we measure the accuracy--latency trade-offs introduced by accelerated decoding and the sensitivity to action discretization and auxiliary-loss scaling.

Unless stated otherwise, the policy receives RGB observations and a natural-language instruction and produces continuous controls by denoising and dequantizing action tokens. Each downstream domain uses a separately adapted checkpoint, allowing us to evaluate the shared modeling and adaptation approach across domains. Task success rate is the primary policy metric. On VLABench, in-distribution (ID) success denotes Track~1 performance, out-of-distribution (OOD) success averages Tracks~3 and~4, and the gap is \(\mathrm{ID}-\mathrm{OOD}\). We interpret this gap together with absolute success, since low performance in both conditions can also produce a small gap.

Cross-model policy tables combine cited and reproduced results obtained with model-specific training recipes; within-model ablations examine the effects of individual design choices or training recipes. We treat numerical differences as descriptive and report variability where available. For checkpoints using the demonstration metadata in Section~\ref{subsubsec:metadata}, rollout prompts contain the specified fixed placeholder values.

\paragraph{LIBERO.}
LIBERO~\cite{liu2023libero} comprises four suites covering spatial relations, object-centric manipulation, goal-conditioned interaction, and long-horizon composition. We report success on each suite and their macro-average to assess standard-task policy performance. Section~\ref{sec:ablation} separately examines how the objective mixture affects downstream behavior. The reported LIBERO result uses the two-stage goal-guided inference path, variant~(c) in Section~\ref{sec:inference_ablation}.

\paragraph{LIBERO-Plus.}
LIBERO-Plus~\cite{fei2025liberoplus} evaluates seven perturbation families: Camera, Robot, Language, Light, Background, Noise, and Layout. We apply the LIBERO-trained checkpoint with the same inference path and no LIBERO-Plus training or adaptation. Here, ``zero-shot'' denotes transfer from LIBERO to these perturbed evaluation conditions. We report each family's success rate and their macro-average to characterize which changes are accommodated by the learned policy and which remain difficult.

\begin{figure}[t]
\centering
\vspace{-0.3em}
\includegraphics[width=1\linewidth]{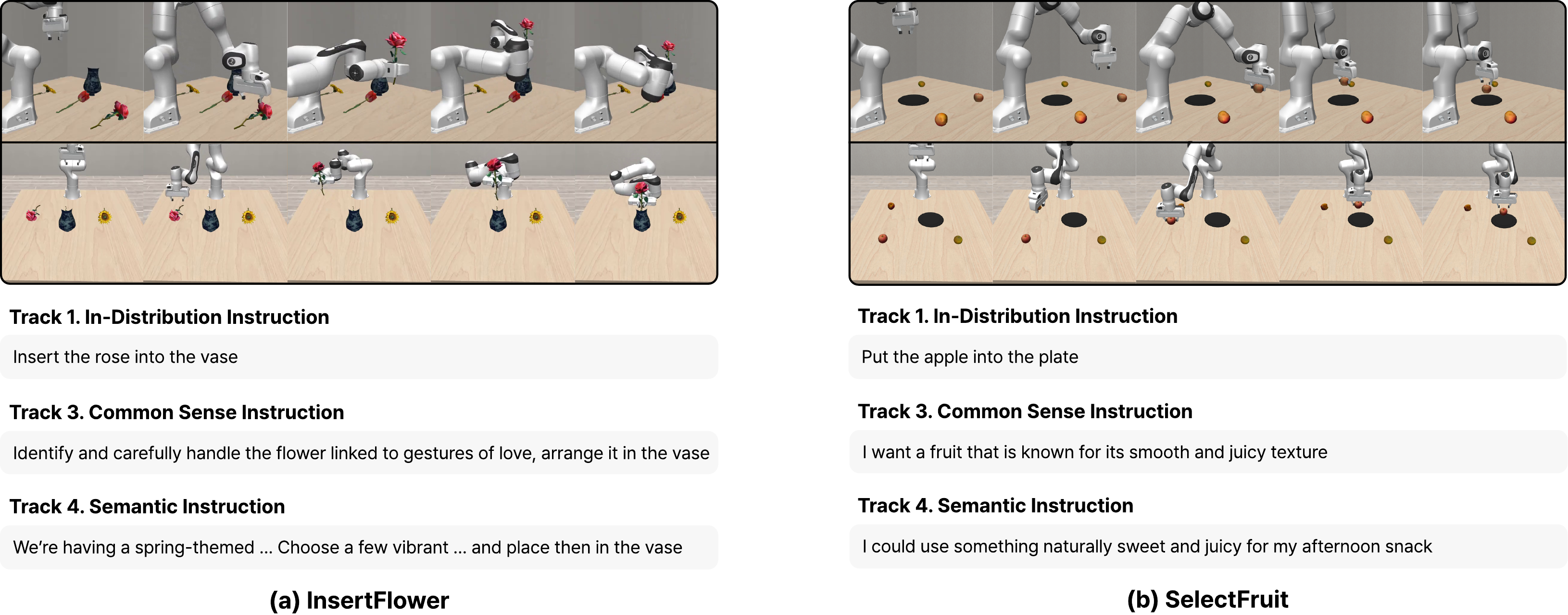}
\vspace{-1.0em}
\caption{VLABench diagnostic tasks and instruction tracks. Example rollouts and language variants are shown for (a) \textsc{InsertFlower}, a contact-rich insertion task requiring precise grasp–alignment–insertion control, and (b) \textsc{SelectFruit}, an object-grounding task requiring adaptation to varying object identities and layouts. Track 1 provides direct in-distribution instructions, Track 3 uses commonsense descriptions, and Track 4 uses indirect semantic paraphrases.}
\label{fig:motivation_example}
\end{figure}

\paragraph{VLABench.}
We study SelectFruit and InsertFlower from VLABench~\cite{zhang2025vlabench}. SelectFruit requires selecting a target across object and layout variation, while InsertFlower requires grasping, alignment, and insertion. Models are trained with direct Track~1 instructions and evaluated with those instructions, commonsense descriptions from Track~3, and indirect semantic paraphrases from Track~4. Figure~\ref{fig:motivation_example} illustrates the two tasks and their instruction formats. The 5k--30k post-training checkpoints in Section~\ref{subsubsec:vlabench_results} characterize learning behavior. The ablations and sensitivity studies in Sections~\ref{sec:ablation} and~\ref{sec:sensitivity} instead compare 70k-step Stage-2 endpoints. These experiments focus on instruction shifts within the two selected tasks.

\paragraph{Real-world experiments.}
We evaluate a Franka Research~3 robot in one workspace under four conditions drawn from three task families. Fruit PnP requires selecting an instructed fruit and placing it at a target. Cube Sort requires placing colored cubes into matching boxes. Cube Stack requires a stable four-cube stack in arbitrary order, while Color-Ordered Stack additionally requires an instruction-specified color sequence. Success is the fraction of rollouts satisfying the corresponding terminal condition, and the overall score is the macro-average across these four conditions. This study assesses execution of the adapted policy on the reported physical setup.

\subsection{Policy Evaluation}
\label{subsec:main_results}

\subsubsection{LIBERO}

\begin{table*}[t]
% \caption{Evaluation results on the LIBERO~\cite{liu2023libero} benchmark. The best score in each suite and the overall average is underlined.}
\caption{Success rates (\%) on the four LIBERO task suites. Methods are grouped into continuous-diffusion, vision-language, video-generation, and unified policy models. Average is the macro-average across the four suites. The best result in each suite and the overall average is underlined.}
\label{tab:libero_result}
\footnotesize
\setlength{\tabcolsep}{14pt}
\renewcommand{\arraystretch}{0.95}
\centering
\begin{tabular}{lccccc}
\toprule
Models & \shortstack{Spatial} & \shortstack{Object} &
\shortstack{Goal} & \shortstack{Long} &
\shortstack{Average}~$\uparrow$ \\
\midrule
\multicolumn{6}{l}{\textit{Continuous Diffusion Models}} \\
Diffusion Policy~\cite{chi2023diffusion}
    & 78.5 & 87.5 & 73.5 & 64.8 & 76.1 \\
\midrule
\multicolumn{6}{l}{\textit{Vision-Language Models}} \\
OpenVLA~\cite{kim2024openvla}
    & 84.7 & 88.4 & 79.2 & 53.7 & 76.5 \\
SpatialVLA~\cite{qu2025spatialvla}
    & 88.2 & 89.9 & 78.6 & 55.5 & 78.1 \\
CoT-VLA~\cite{zhao2025cot}
    & 87.5 & 91.6 & 87.6 & 69.0 & 83.9 \\
$\pi_0$-FAST~\cite{pertsch2025fast}
    & 96.4 & 96.8 & 88.6 & 60.2 & 85.5 \\
GR00T-N1~\cite{bjorck2025gr00t}
    & 94.4 & 97.6 & 93.0 & 90.6 & 93.9 \\
$\pi_0$~\cite{black2024pi0}
    & 98.0 & 96.8 & 94.4 & 88.4 & 94.4 \\
F1~\cite{lv2025f1}
    & 98.2 & 97.8 & 95.4 & 91.3 & 95.7 \\
InternVLA-M1~\cite{internvlam1}
    & 98.0 & 99.0 & 93.8 & 92.6 & 95.9 \\
Discrete Diffusion VLA~\cite{liang2025discretediffusionvla}
    & 97.2 & 98.6 & 97.4 & 92.0 & 96.3 \\
$\pi_{0.5}$~\cite{intelligence2025pi05}
    & 98.8 & 98.2 & 98.0 & 92.4 & 96.9 \\
GR00T-N1.6~\cite{bjorck2025gr00t}
    & 97.7 & 98.5 & 97.5 & 94.4 & 97.0 \\
OpenVLA-OFT~\cite{kim2025openvlaoft}
    & 97.6 & 98.4 & 97.9 & 94.5 & 97.1 \\
X-VLA~\cite{zheng2025xvla}
    & 98.2 & 98.6 & 97.8 & 97.6 & 98.1 \\
ABot-M0~\cite{yang2026abot}
    & 98.8 & 99.8 & \underline{99.0} & 96.6 & \underline{98.6} \\
\midrule
\multicolumn{6}{l}{\textit{Video Generation Models}} \\
Video Policy~\cite{liang2025videopolicy}
    & - & - & - & 94.0 & - \\
Cosmos Policy~\cite{kim2026cosmos}
    & 98.1 & \underline{100.0} & 98.2 & 97.6 & 98.5 \\
CogVLA~\cite{li2025cogvla}
    & 98.6 & 98.8 & 96.6 & 95.4 & 97.4 \\
Mimic-Video~\cite{pai2025mimic}
    & 96.2 & 98.3 & 96.2 & - & - \\
LingBot-VA~\cite{li2026lingbotva}
    & 98.5 & 99.6 & 97.2 & \underline{98.5} & 98.5 \\
Fast-WAM~\cite{yuan2026fastwam}
    & 98.2 & \underline{100.0} & 97.0 & 95.2 & 97.6 \\
\midrule
\multicolumn{6}{l}{\textit{Unified Models}} \\
dVLA~\cite{wen2025dvla}
    & 97.4 & 97.9 & 98.2 & 92.2 & 96.4 \\
WorldVLA~\cite{cen2025worldvla}
    & 87.6 & 96.2 & 83.4 & 60.0 & 81.8 \\
RynnVLA-002~\cite{cen2025rynnvla002}
    & \underline{99.0} & 99.8 & 96.4 & 94.4 & 97.4 \\
UD-VLA~\cite{chen2025udvla}
    & 94.1 & 95.7 & 91.2 & 89.6 & 92.7 \\
MMaDA-VLA~\cite{liu2026mmada}
    & 98.8 & 99.8 & 98.0 & 95.2 & 98.0 \\
\rowcolor{lightblue}
% Dynin-Robotics (Ours)
%     & 98.9 $\pm$ 0.3 & 99.8 $\pm$ 0.1 & 97.8 $\pm$ 0.4 & 95.8 $\pm$ 0.5 & 98.1 \\
Dynin-Robotics (Ours)
    & 98.9 & 99.8 & 97.8 & 95.8 & 98.1 \\
\bottomrule
\end{tabular}
\end{table*}

Dynin-Robotics achieves a 98.1\% macro-average on LIBERO (Table~\ref{tab:libero_result}), matching X-VLA and placing 0.5 percentage points below ABot-M0. Several methods obtain closely grouped scores near saturation, positioning Dynin-Robotics among the stronger policies in this comparison.

The suite profile reveals where performance remains uneven. Dynin-Robotics reaches 98.9\% on Spatial and 99.8\% on Object, compared with 97.8\% on Goal and 95.8\% on Long. Long-horizon tasks retain the most room for improvement among the four suites. This performance establishes that the shared trajectory model supports high policy success alongside its visual and language output interfaces.

\subsubsection{LIBERO-Plus}

\begin{table*}[t]
% \caption{
% Zero-shot success rates (\%) under the seven LIBERO-Plus perturbation categories. Models are evaluated on camera, robot, language, lighting, background, noise, and layout perturbations without task-specific adaptation. Average denotes the macro-average across perturbation categories. The best result in each category and the overall average is underlined.
% }
\caption{Zero-shot transfer from LIBERO to the seven LIBERO-Plus perturbation categories without additional LIBERO-Plus training. Entries report success rates (\%) under camera, robot, language, lighting, background, noise, and layout perturbations. Average is recomputed as the unweighted mean of the seven displayed category scores. Underlining marks the highest value in each column.}
\label{tab:libero_plus_result}
\footnotesize
\setlength{\tabcolsep}{5.5pt}
\renewcommand{\arraystretch}{0.95}
\centering
\begin{tabular}{lcccccccc}
\toprule
Models
& \shortstack{Camera}
& \shortstack{Robot}
& \shortstack{Language}
& \shortstack{Light}
& \shortstack{Background}
& \shortstack{Noise}
& \shortstack{Layout}
& \shortstack{Average}~$\uparrow$ \\
\midrule
% \multicolumn{9}{l}{\textit{Vision-Language Model}} \\
OpenVLA~\cite{kim2024openvla}
& 0.8
& 3.5
& 23.0
& 8.1
& 34.8
& 15.2
& 28.5
& 16.3 \\

OpenVLA-OFT~\cite{kim2025openvlaoft}
& 56.4
& 31.9
& 79.5
& 88.7
& 93.3
& 75.8
& 74.2
& 71.4 \\

OpenVLA-OFT\_w~\cite{kim2025openvlaoft}
& 10.4
& 38.7
& 70.5
& 76.8
& \underline{93.6}
& 49.9
& 69.9
& 58.5 \\

OpenVLA-OFT\_m~\cite{kim2025openvlaoft}
& 55.6
& 21.7
& 81.0
& 92.7
& 91.0
& 78.6
& 68.7
& 69.9 \\

NORA~\cite{hung2025nora}
& 2.2
& 37.0
& 65.1
& 45.7
& 58.6
& 12.8
& 62.1
& 40.5 \\

% WorldVLA~\cite{cen2025worldvla}
% & 0.1
% & 27.9
% & 41.6
% & 43.7
% & 17.1
% & 10.9
% & 38.0
% & 25.0 \\

UniVLA~\cite{wang2025unifiedvla}
& 1.8
& 46.2
& 69.6
& 69.0
& 81.0
& 21.2
& 31.9
& 45.8 \\

$\pi_0$~\cite{black2024pi0}
& 13.8
& 6.0
& 58.8
& 85.0
& 81.4
& 79.0
& 68.9
& 56.1 \\

$\pi_0$-FAST~\cite{pertsch2025fast}
& \underline{65.1}
& 21.6
& 61.0
& 73.2
& 73.2
& 74.4
& 68.8
& 62.5 \\

RIPT-VLA~\cite{tan2025ript}
& 55.2
& 31.2
& 77.6
& 88.4
& 91.6
& 73.5
& 74.2
& 70.2 \\

ABot-M0~\cite{yang2026abot}
& 60.4
& \underline{67.9}
& \underline{86.4}
& \underline{96.2}
& 91.6
& \underline{86.4}
& \underline{82.6}
& \underline{81.6} \\

% \midrule
% \multicolumn{9}{l}{\textit{Video Model}} \\
% - & . & . & . & . & . & . & . & . \\
% \midrule
% \multicolumn{9}{l}{\textit{Unified Model}} \\
WorldVLA~\cite{cen2025worldvla}
& 0.1
& 27.9
& 41.6
& 43.7
& 17.1
& 10.9
& 38.0
& 25.6 \\

\rowcolor{lightblue}
Dynin-Robotics (Ours)
& 59.8 & 48.2 & 85.0 & 83.5 & 84.6 & 78.2 & 71.8 & 73.0 \\
% Dynin-Robotics (Ours)
% & 59.8 $\pm$ 1.3
% & 48.2 $\pm$ 1.9
% & 85.0 $\pm$ 0.9
% & 83.5 $\pm$ 2.1
% & 84.6 $\pm$ 0.5
% & 78.2 $\pm$ 1.4
% & 71.8 $\pm$ 1.0
% & 73.0 \\
\bottomrule
\end{tabular}
\end{table*}

Transfer to perturbed conditions reveals a less uniform performance profile. Dynin-Robotics obtains a 73.0\% macro-average in Table~\ref{tab:libero_plus_result}, above the listed OpenVLA-OFT and RIPT-VLA results and below ABot-M0 at 81.6\%. Language perturbations retain an 85.0\% success rate, whereas Camera and Robot yield 59.8\% and 48.2\%, respectively. The high clean-benchmark score therefore coexists with substantial sensitivity to some of the evaluated changes.

The perturbation families also contribute unevenly to the difference from ABot-M0. Camera is one of Dynin-Robotics' lowest-scoring categories, yet its gap to ABot-M0 is only 0.6 percentage points. The gaps are larger for Robot and Light, at 19.7 and 12.7 points. The family-level results thus distinguish conditions that are difficult for both models from those where Dynin-Robotics falls further behind, identifying priorities for improving transfer.

\subsubsection{VLABench}
\label{subsubsec:vlabench_results}

\begin{figure}[t]
\centering
\vspace{-0.3em}
\includegraphics[width=1\linewidth]{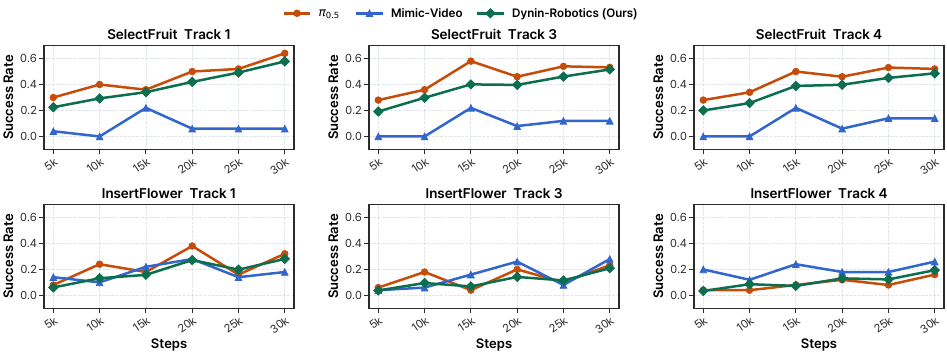}
\vspace{-1.0em}
\caption{Intermediate VLABench checkpoint performance from 5k to 30k post-training steps for $\pi_{0.5}$, Mimic-Video, and Dynin-Robotics. The top row shows SelectFruit and the bottom row InsertFlower; curves compare direct, commonsense, and indirect instructions from Tracks~1, 3, and~4. These diagnostics are distinct from the 70k endpoints used in the ablation and sensitivity studies.}
\label{fig:motivation_result}
\end{figure}

\begin{table*}[!htbp]
\centering
\caption{Instruction sensitivity on VLABench. Success rate and progress score are measured with the original instruction and a length-matched random string under the same visual observation and rollout settings. Parentheses show the signed change (random minus original).}
\label{tab:random_instruction}
\footnotesize
\setlength{\tabcolsep}{16pt}
\renewcommand{\arraystretch}{0.95}
\begin{tabular}{@{}llcccc@{}}
\toprule
& & \multicolumn{2}{c}{Success Rate~$\uparrow$} &
\multicolumn{2}{c}{Progress Score~$\uparrow$} \\
\cmidrule(lr){3-4}\cmidrule(lr){5-6}
Model & Task & Original & Random &
Original & Random \\
\midrule
Mimic-Video~\cite{pai2025mimic} & \textsc{InsertFlower}
& 0.18 & 0.28\,{\tiny\textcolor{green!50!black}{(+0.10)}}
& 0.58 & 0.63\,{\tiny\textcolor{green!50!black}{(+0.05)}} \\
& \textsc{SelectFruit}
& 0.08 & 0.10\,{\tiny\textcolor{green!50!black}{(+0.02)}}
& 0.53 & 0.53\,{\tiny(0.00)} \\
\midrule
$\pi_{0.5}$~\cite{intelligence2025pi05} & \textsc{InsertFlower}
& 0.32 & 0.16\,{\tiny\textcolor{red!75!black}{(-0.16)}}
& 0.65 & 0.57\,{\tiny\textcolor{red!75!black}{(-0.08)}} \\
& \textsc{SelectFruit}
& 0.64 & 0.34\,{\tiny\textcolor{red!75!black}{(-0.30)}}
& 0.82 & 0.53\,{\tiny\textcolor{red!75!black}{(-0.29)}} \\
\bottomrule
\end{tabular}
\end{table*}

The two-task diagnostic examines how policy performance varies with the manipulation task and instruction form. In Figure~\ref{fig:motivation_result}, \(\pi_{0.5}\) performs better than Mimic-Video on SelectFruit across the displayed tracks. On InsertFlower, Mimic-Video is particularly competitive under the indirect Track~4 instructions, while the direct-instruction curves show a different ordering. Dynin-Robotics substantially exceeds Mimic-Video on SelectFruit and remains competitive on InsertFlower, with its relative position varying across checkpoints and tracks. These observations reveal distinct task and instruction profiles among the evaluated policies.

The curves track performance across post-training checkpoints. Together with the absolute success rates, they show why a small separation between instruction tracks is insufficient to characterize robustness: a policy can perform similarly across tracks while solving few tasks. The 70k-step ablations below therefore compare absolute ID and OOD success alongside their gap.

Instruction randomization further distinguishes the two reference policies. In Table~\ref{tab:random_instruction}, replacing the instruction with a length-matched random string reduces \(\pi_{0.5}\)'s success by 16 and 30 percentage points on InsertFlower and SelectFruit. Mimic-Video instead changes by $+10$ and $+2$ points, with progress scores showing the same qualitative contrast. Thus, \(\pi_{0.5}\) is more sensitive to this text intervention on the two tasks, while Mimic-Video retains its success under instruction randomization. Together with the checkpoint profiles, this observation motivates evaluating task success and instruction sensitivity as complementary aspects of policy behavior.

\subsubsection{Real-World Manipulation}

\begin{figure}[t]
\centering
\vspace{-0.3em}
\includegraphics[width=1\linewidth]{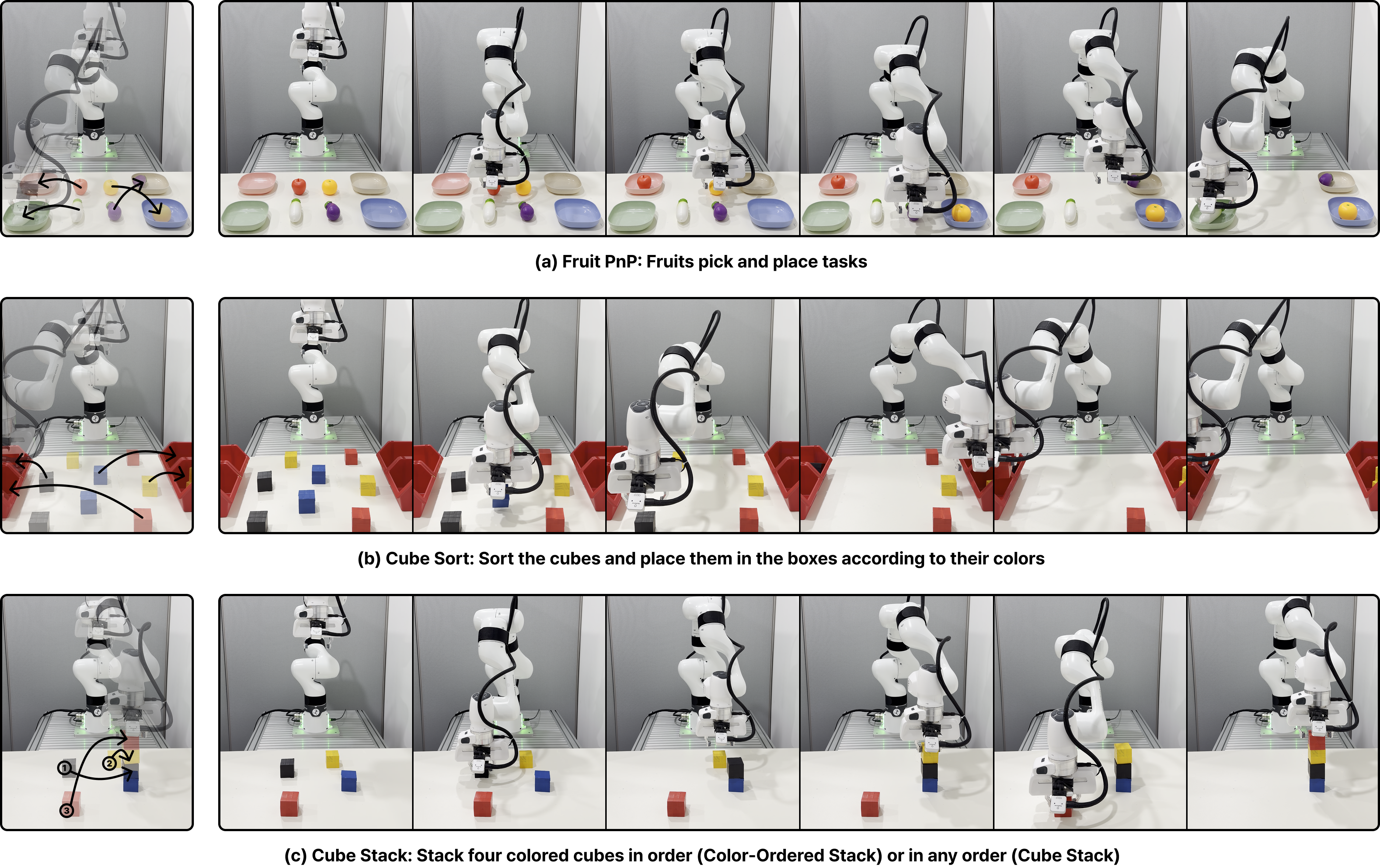}
\vspace{-1.0em}
% \caption{
% Real-world manipulation tasks on the Franka Research 3 platform,
% including fruit pick-and-place (\textsc{Fruit PnP}), sorting tabletop cubes into
% color-matched boxes (\textsc{Cube Sort}), stacking cubes in any order (\textsc{Cube Stack}),
% and stacking cubes according to a specified color sequence
% (\textsc{Color-Ordered Stack}).
% }
\caption{
Real-world manipulation tasks on the Franka Research 3 platform. (a) \textsc{Fruit PnP} selects an instructed fruit and places it at the target location. (b) \textsc{Cube Sort} places colored cubes into color-matched boxes. (c) \textsc{Cube Stack} builds a four-cube stack either in arbitrary order or according to an instruction-specified color sequence, denoted \textsc{Color-Ordered Stack}.
}
\label{fig:real_world_task}
\end{figure}

\begin{table*}[!htbp]
\centering
% \caption{Real-world success rates on one Franka Research~3 robot. Average is the macro-average across the four evaluation conditions; entries are point estimates without uncertainty intervals. The best result in each condition and the overall average is underlined.}
\caption{Real-world manipulation success rates (\%) on a Franka Research~3 robot. Fruit PnP, Cube Sort, Cube Stack, and Color-Ordered Stack define four evaluation conditions, with the latter two differing in whether the stacking order is specified by the instruction. Average is the unweighted mean across the four conditions. Underlining marks the highest value in each column.}
\label{tab:realworld}
\footnotesize
\setlength{\tabcolsep}{8.9pt}
\renewcommand{\arraystretch}{0.95}
\begin{tabular}{lccccc}
\toprule
Method & \textsc{Fruit PnP} & \textsc{Cube Sort} &
\textsc{Cube Stack} & \textsc{Color-Ordered Stack} &
Average(\%)~$\uparrow$ \\
\midrule
\multicolumn{6}{l}{\textit{Vision-Language Models}} \\
$\pi_{0.5}$~\cite{intelligence2025pi05} & 96.5 & 83.5 & 63.5 & 64.0 & 76.9 \\
GR00T-N1.6~\cite{bjorck2025gr00t} & 91.0 & 78.0 & 59.5 & 63.5 & 73.0 \\
\midrule
\multicolumn{6}{l}{\textit{Video-Generation Models}} \\
Cosmos Policy~\cite{kim2026cosmos} & \underline{98.0} & \underline{87.0} & \underline{67.5} & 42.0 & 73.6 \\
Mimic-Video~\cite{pai2025mimic} & 96.5 & 84.5 & 64.0 & 39.5 & 71.1 \\
\midrule
\multicolumn{6}{l}{\textit{Unified Models}} \\
\rowcolor{lightblue}
Dynin-Robotics (Ours) & 97.5 & 85.5 & 62.0 & \underline{68.5} & \underline{78.4} \\
\bottomrule
\end{tabular}
\end{table*}

The physical evaluation covers object selection, sorting, and stacking on the Franka Research~3 platform. Figure~\ref{fig:real_world_task} illustrates the task families, including the additional order constraint in Color-Ordered Stack. The four conditions allow us to examine how task requirements affect the comparative performance of the adapted policies.

Dynin-Robotics achieves the highest reported macro-average, 78.4\%, and the highest Color-Ordered Stack success, 68.5\%, in Table~\ref{tab:realworld}. Relative to \(\pi_{0.5}\), its condition-wise differences are $+1.0$, $+2.0$, $-1.5$, and $+4.5$ percentage points for Fruit PnP, Cube Sort, Cube Stack, and Color-Ordered Stack, respectively. Its 1.5-point average advantage is therefore unevenly distributed, with the largest positive difference occurring when the stacking order is specified by language.

Cosmos Policy obtains the strongest results on Fruit PnP, Cube Sort, and unconstrained Cube Stack, whereas Dynin-Robotics leads on Color-Ordered Stack. The condition-level results therefore reveal different strengths beneath the overall ranking. Dynin-Robotics' clearest comparative advantage occurs in stacking with an instruction-specified order, a task that combines object manipulation with a sequence constraint.

\subsection{World Modeling and Goal-State Prediction}
\label{subsec:world_model_performance}

Direct output evaluation examines the predictions provided by the shared trajectory interface. World Modeling predicts the immediately subsequent visual state from the current context and corresponding action; Goal-State Prediction generates a terminal task state from the initial observation and instruction. All quantitative and qualitative Dynin-Robotics outputs in this section use the Stage-1 checkpoint on the DROID validation set without additional DROID-specific post-training. DROID~\cite{khazatsky2024droid} is included in the Stage-1 OXE mixture.

We measure reference-image similarity using peak signal-to-noise ratio (PSNR), structural similarity (SSIM), and learned perceptual image patch similarity (LPIPS); higher PSNR and SSIM and lower LPIPS indicate better agreement. These metrics characterize image reconstruction, while the inference ablations evaluate the use of predictions during control. Their interpretation depends on the prediction task: adjacent frames can share substantial static content, and distinct terminal configurations can satisfy the same instruction. Cross-model rows use each model's reported prediction interface, providing a comparison of output quality across those interfaces.

\subsubsection{Comparison with Unified Policies}

\definecolor{adaptedcell}{RGB}{255,246,210}
\definecolor{nacell}{RGB}{235,235,235}
\definecolor{lightblue}{RGB}{221,235,247}

\newcommand{\adapted}{\cellcolor{adaptedcell}--}
\newcommand{\na}{\cellcolor{nacell}--}

\begin{table}[t]
\centering
% \caption{
% Evaluation of world modeling and policy prediction on DROID.
% World Modeling predicts adjacent future observations from current
% observations, optionally conditioned on robot actions or language instructions.
% Policy evaluates coordinate-level action accuracy at a normalized tolerance
% of 0.1 and normalized continuous action error.
% Gray cells indicate unsupported objectives.
% Higher is better ($\uparrow$), except LPIPS and Action MAE ($\downarrow$). The best score in each metric is underlined except heuristic-based (Last Frame and Random Frame).
% }
% \caption{World-modeling and policy-prediction performance on DROID. World Modeling predicts adjacent future observations; Policy predicts normalized actions. Image quality uses PSNR, SSIM, and LPIPS, and action quality uses accuracy at tolerance 0.1 and normalized mean absolute error. Gray cells denote unsupported outputs; best-score underlining excludes the Last Frame and Random Frame heuristics. The Dynin-Robotics row uses the Stage-1 checkpoint evaluated zero-shot on the DROID validation set.}
\caption{Visual and offline action prediction on the DROID validation set. Adjacent-future image predictions are evaluated with PSNR, SSIM, and LPIPS, and normalized action predictions with accuracy at tolerance 0.1 and mean absolute error. Models use their respective prediction interfaces; Dynin-Robotics uses action-conditioned World Modeling. Its results are obtained from the Stage-1 checkpoint without additional DROID-specific post-training, with DROID included in the Stage-1 mixture. Underlining marks the best values among learned models, excluding the Last Frame and Random Frame references. Dashes indicate metrics not reported for the corresponding row.}
\label{tab:world_policy_benchmark}
\footnotesize
\setlength{\tabcolsep}{14pt}
\renewcommand{\arraystretch}{1.05}

\begin{tabular}{lccccc}
\toprule
&
\multicolumn{3}{c}{World Modeling}
&
\multicolumn{2}{c}{Policy}
\\

\cmidrule(lr){2-4}
\cmidrule(l){5-6}

Model
& PSNR~$\uparrow$
& SSIM~$\uparrow$
& LPIPS~$\downarrow$
& Action Acc$@$0.1~$\uparrow$
& Action MAE~$\downarrow$
\\

\midrule

\textit{Baselines} & & & & & \\

Last Frame
& 31.07
& 0.953
& 0.013
& --
& --
\\

Random Frame
& 8.94
& 0.155
& 0.665
& --
& --
\\

\midrule

\textit{Unified VLA} & & & & & \\

UP-VLA
& 11.19
& 0.290
& 0.491
& 0.2952
& 0.2492
\\

MMaDA-VLA
& 17.72
& 0.533
& \underline{0.232}
& 0.3204
& 0.3068
\\

\rowcolor{lightblue}
Dynin-Robotics (Ours)
& \underline{20.79}
& \underline{0.573}
& 0.299
& \underline{0.3327}
& \underline{0.2109}
\\

\bottomrule
\end{tabular}
\end{table}

\begin{figure}[ht]
\centering
\vspace{-0.3em}
\includegraphics[width=1.0\linewidth]{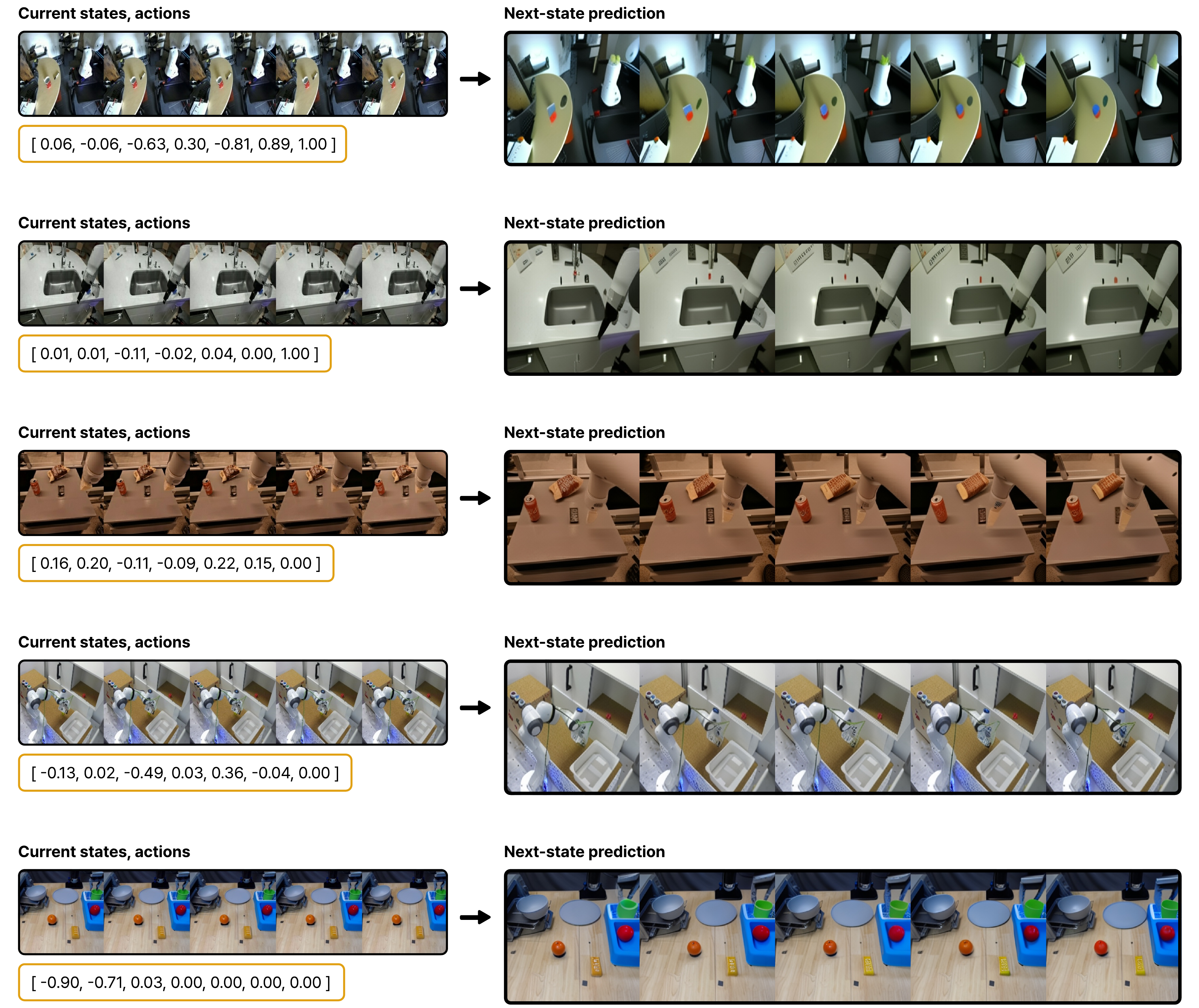}
% \caption{Qualitative World Modeling outputs from the Stage-1 checkpoint evaluated zero-shot on the DROID validation set. Each example shows current-state frames and the conditioning action chunk (left) and Dynin-Robotics' predicted adjacent-future frames (right). Ground-truth next frames are not displayed, so the figure illustrates the output interface rather than a paired fidelity comparison.}
\caption{Qualitative action-conditioned next-observation prediction on the DROID validation set using the Stage-1 checkpoint without additional DROID-specific post-training. Each example shows the visual observations and action context on the left and the predicted adjacent-future observations on the right.}
\label{fig:qualitative_result_wm}
\end{figure}

Frame copying provides a strong reference for adjacent-future prediction in Table~\ref{tab:world_policy_benchmark}. Copying the last frame yields 31.07 PSNR, 0.953 SSIM, and 0.013 LPIPS, compared with Dynin-Robotics at 20.79, 0.573, and 0.299. The generated predictions therefore fall below persistence on all three whole-image metrics. Because these scores aggregate static scene content and changing regions, evaluating action-dependent changes remains a distinct requirement for assessing the learned transition model.

Among the learned predictors in the table, Dynin-Robotics has the highest PSNR and SSIM, while MMaDA-VLA has the lowest LPIPS. The same Dynin-Robotics checkpoint also achieves the highest reported offline action accuracy at tolerance 0.1 (0.3327) and the lowest normalized action MAE (0.2109). These results show complementary strengths across output metrics within a model that supports both visual and action prediction.

Figure~\ref{fig:qualitative_result_wm} illustrates next-state outputs generated from robot observations and action conditioning across several scenes. The examples show the visual content produced by this interface; the quantitative comparison above measures agreement with recorded futures. Section~\ref{sec:inference_ablation} evaluates how composing these predictions with action generation affects policy success.

\subsubsection{Comparison with Image-Generation Models}

\begin{table*}[t]
\centering
% \caption{
% Evaluation of multimodal robotics objectives on DROID.
% World Modeling predicts adjacent future observations from current observations.
% Task Understanding predicts the task instruction from trajectory observations.
% Goal-State Prediction predicts the terminal successful state from the
% initial observation and instruction.
% Unmarked numeric cells use the model's native interface.
% Higher is better ($\uparrow$), except LPIPS ($\downarrow$).
% }
% \caption{World-modeling and goal-state image quality on DROID. World Modeling predicts adjacent future observations; Goal-State Prediction generates terminal successful states from initial observations and instructions. Both use PSNR, SSIM, and LPIPS; arrows indicate metric direction, and underlining marks the best value in each column. The Dynin-Robotics row uses the Stage-1 checkpoint evaluated zero-shot on the DROID validation set.}
\caption{Visual prediction quality on the DROID validation set for adjacent-future observations and terminal goal states. Predictions are evaluated with PSNR, SSIM, and LPIPS using each model's prediction interface. Dynin-Robotics uses the Stage-1 checkpoint without additional DROID-specific post-training. Arrows indicate metric direction, and underlining marks the best value in each column.}
\label{tab:goal_benchmark}
\footnotesize
\setlength{\tabcolsep}{13pt}
\renewcommand{\arraystretch}{1.05}

\begin{tabular}{lcccccc}
\toprule
&
\multicolumn{3}{c}{World Modeling}
% &
% \multicolumn{2}{c|}{Task Understanding}
&
\multicolumn{3}{c}{Goal-State Prediction}
\\

\cmidrule(lr){2-4}
% \cmidrule(lr){5-6}
\cmidrule(lr){5-7}
% \cmidrule(l){7-9}

Model
& PSNR~$\uparrow$
& SSIM~$\uparrow$
& LPIPS~$\downarrow$
% & Acc.$\uparrow$
% & BLEU-4$\uparrow$
& PSNR~$\uparrow$
& SSIM~$\uparrow$
& LPIPS~$\downarrow$
\\

\midrule

BAGEL
& 19.3063
& \underline{0.6405}
& \underline{0.1568}
% & 0.0074$^\dagger$
% & 0.0617$^\dagger$
& \underline{13.4755}
& \underline{0.5005}
& \underline{0.3626}
\\

MMaDA
& 8.8326
& 0.2120
& 0.7532
% & 0.0000
% & 0.00002
& 8.6710
& 0.2058
& 0.7630
\\

Show-o2
& 8.7519
& 0.2098
& 0.7220
% & 0.0000$^\dagger$
% & 0.0084$^\dagger$
& 8.7544
& 0.1850
& 0.7275
\\

% Dynin-Omni
% & 11.3358
% & 0.2475
% & 0.4837
% & \cellcolor{adaptedcell}0.0000$^\dagger$
% & \cellcolor{adaptedcell}0.0109$^\dagger$
% & 9.5831
% & 0.2319
% & 0.5791
% \\

\rowcolor{lightblue}
Dynin-Robotics (Ours)
& \underline{20.7941}
& 0.5731
& 0.2994
% & --
% & --
& 10.7098
& 0.2825
& 0.4715
\\

\bottomrule
\end{tabular}
\end{table*}

\begin{figure}[ht]
\centering
\vspace{-0.3em}
\includegraphics[width=1.0\linewidth]{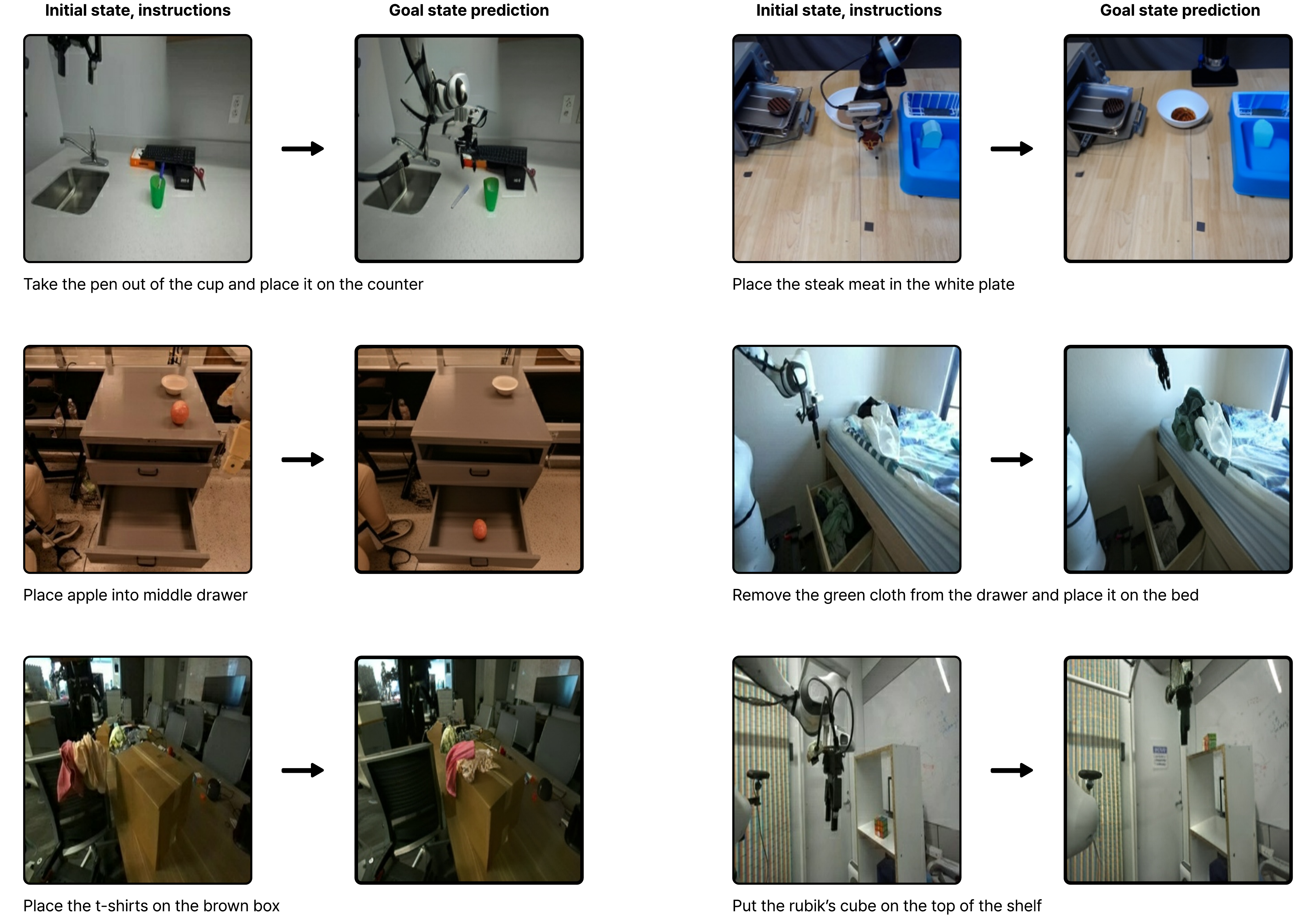}
% \caption{Qualitative Goal-State Prediction outputs from the Stage-1 checkpoint evaluated zero-shot on the DROID validation set. Each row shows the initial observation and task instruction (left) and Dynin-Robotics' generated terminal state (right). Ground-truth terminal images are not displayed.}
\caption{Qualitative Goal-State Prediction on the DROID validation set using the Stage-1 checkpoint without additional DROID-specific post-training. Each example pairs an initial observation and task instruction with the predicted terminal visual state, shown on the right.}
\label{fig:qualitative_result_goal}
\end{figure}

Comparison with image-generation models shows that the relative performance depends on the prediction task and metric. In Table~\ref{tab:goal_benchmark}, Dynin-Robotics has the highest World Modeling PSNR, whereas BAGEL has higher SSIM and lower LPIPS. For Goal-State Prediction, BAGEL leads on all three metrics and Dynin-Robotics ranks second among the listed models. Dynin-Robotics is therefore strongest in adjacent-future pixel reconstruction within this comparison, with greater room for improvement in terminal-state generation.

The two prediction settings also play different roles during control: the next-state interface represents a local transition, while the goal interface supplies a visual target for task completion. For goal prediction, reference-image agreement captures only one aspect of output quality, since alternative successful terminal configurations can differ from the recorded image. Task-condition satisfaction and the effect on policy success provide complementary criteria for evaluating generated goals.

Figure~\ref{fig:qualitative_result_goal} shows selected initial observations, instructions, and generated goal images for tasks involving object placement and scene rearrangement. These examples illustrate the terminal-state predictions available for goal conditioning. The inference-composition study evaluates their use by comparing goal-conditioned and action-only decoding on the same checkpoint.

% \subsection{Qualitative Task Understanding}

% \input{figures/tex/qualitative_results_und}

% Figure~\ref{fig:qualitative_result_und} presents qualitative Task Understanding results. Given sampled visual states from a robot trajectory, Dynin-Robotics generates a natural-language instruction describing the demonstrated task. These examples illustrate the model's ability to infer task semantics from observed trajectories. The contribution of Task Understanding supervision to downstream policy performance is further evaluated through the objective ablation in Section~\ref{sec:ablation}.

\subsection{Qualitative Task Understanding}

\begin{figure}[ht!]
\centering
\vspace{-0.3em}
\includegraphics[width=0.96\linewidth]{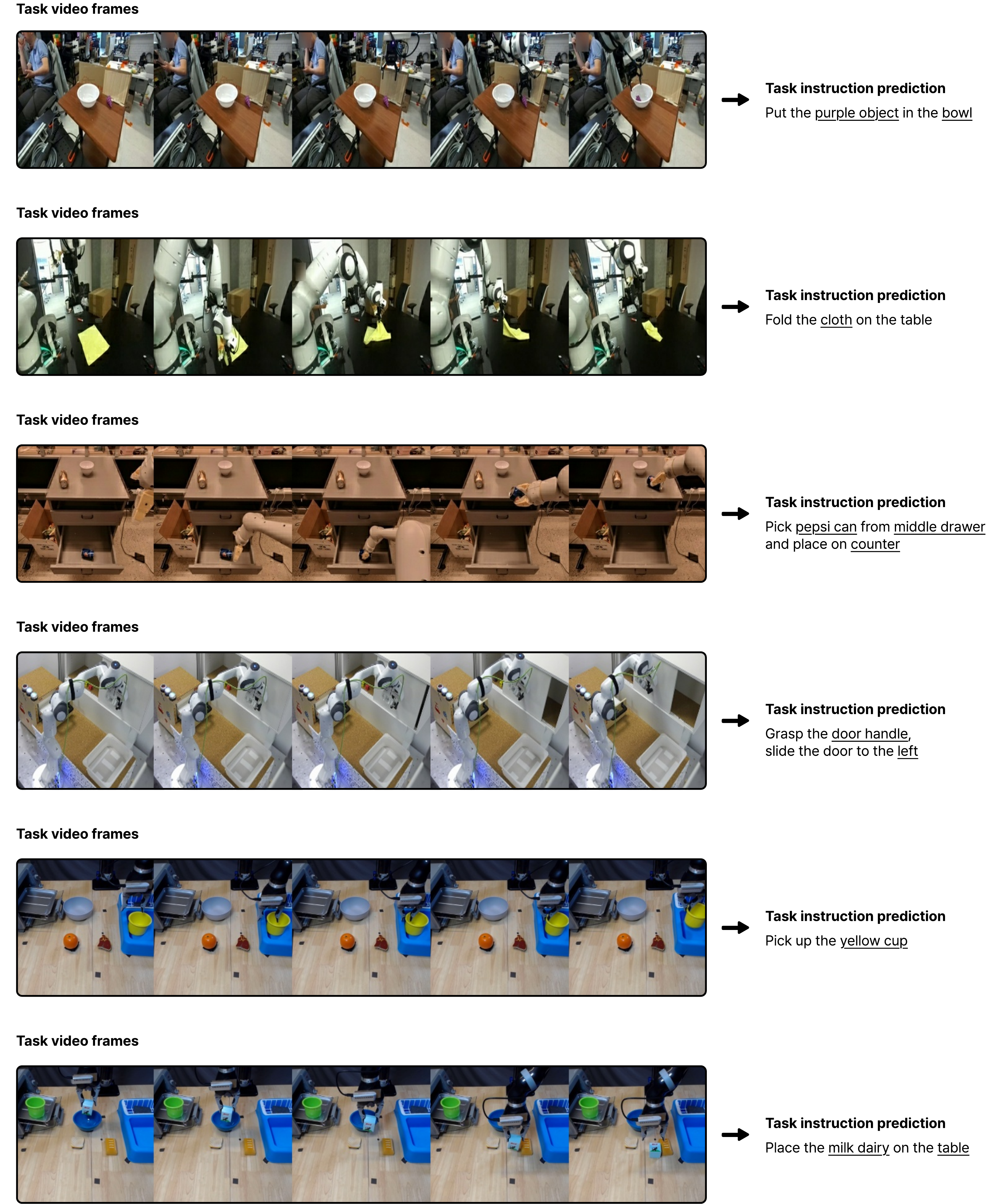}
% \caption{Qualitative Task Understanding outputs from the Stage-1 checkpoint evaluated zero-shot on the DROID validation set. Given sampled trajectory frames, Dynin-Robotics predicts a task instruction. Reference instructions are not displayed, so the figure does not by itself establish semantic accuracy.}
\caption{Qualitative Task Understanding on the DROID validation set using the Stage-1 checkpoint without additional DROID-specific post-training. Sampled trajectory frames are shown on the left, with the generated task descriptions on the right. Underlining highlights object references and spatial expressions in the model outputs.}
\label{fig:qualitative_result_und}
\end{figure}

Task Understanding queries the shared backbone in the trajectory-to-language direction: sampled trajectory frames provide context for generating a task description. Figure~\ref{fig:qualitative_result_und} presents selected outputs from the Stage-1 checkpoint on DROID validation trajectories without additional DROID-specific post-training. The descriptions express action verbs such as picking, placing, folding, and sliding; object attributes such as color; and spatial relations involving bowls, drawers, and counters. One example combines retrieval from a drawer with placement on a counter, describing a sequence that spans multiple local actions.

The outputs vary in specificity, from referring to a purple object to naming a can and its source and destination. This variation illustrates the granularity of the generated descriptions. Quantitative evaluation against reference instructions remains a next step for assessing the accuracy of trajectory-to-language prediction.

The cumulative objective study in Section~\ref{sec:ablation} evaluates Task Understanding from the perspective of policy adaptation. Adding it to the Stage-2 mixture accompanies higher downstream success under the coupled decoder, connecting trajectory-language supervision to the performance of the overall training recipe.

\subsection{Acceleration}
\label{subsec:evaluation_acceleration}

We evaluate the throughput and prediction fidelity of the accelerated action-decoding implementation built on dInfer~\cite{ma2025dinferefficientinferenceframework}. The measured implementation combines modality-aware denoising, confidence-based block commitment, compiled execution, CUDA Graph replay, and loop unrolling. Profiling uses NVIDIA B200 hardware, batch size~1, a 2,048-token context, and a 35-token action chunk corresponding to five 7-DoF actions. Internal Dynin-Robotics fidelity measurements use a commitment threshold of 0.9.

Effective TPS measures action-output throughput in 7-DoF action-token units, while query latency measures generation of a complete five-action chunk. These are model-side action-decoding measurements. Goal generation, sensing, communication, environment stepping, and the execution and replanning schedule are outside their scope. BL7 and BL35 identify the configured token-block lengths; the realized number of backbone evaluations per chunk is measured separately below. For continuous-output baselines, effective TPS provides a common action-output unit rather than a literal count of discrete tokens emitted by their native interfaces.

\begin{table}[!htbp]
\centering
% \caption{Action-decoding throughput under the common one-B200 profiling protocol. Effective TPS counts generated 7-DoF action tokens per second; parentheses show speedup over unaccelerated Dynin-Robotics, and the highest throughput is underlined.}
\caption{Model-side action-decoding throughput under the common single-B200 profiling protocol. Effective TPS reports scalar action-output units per second, with seven units per action. For discrete-token decoders, TPF is the total number of finalized action tokens divided by the total number of backbone evaluations; dashes indicate that this measure is not applicable to the continuous-output interface. Dynin-Robotics TPF values are derived from the reported mean evaluations per 35-token chunk in Table~\ref{tab:dinfer_control_tradeoff}. BL7 and BL35 denote configured block lengths. Parentheses report speedup over the base Dynin-Robotics decoder. Timing excludes goal generation, sensing, communication, and robot execution. The highest throughput is underlined.}
\label{tab:throughput}
\footnotesize
\setlength{\tabcolsep}{16pt}
o\renewcommand{\arraystretch}{0.95}
\begin{tabular}{lcc}
\toprule  
Model & Effective TPS~$\uparrow$ & TPF \\
\midrule
\multicolumn{3}{l}{\textit{Vision-Language Models}} \\
OpenVLA-OFT~\cite{kim2025openvlaoft} & 19.114 & -- \\
$\pi_{0.5}$~\cite{intelligence2025pi05} & 7.351 & -- \\
\midrule
\multicolumn{3}{l}{\textit{Masked-Diffusion Models}} \\
LLaDA-VLA~\cite{wen2025llada} & 2.079 & 1.00 \\
MMaDA-VLA~\cite{liu2026mmada} & 1.827 & 1.00 \\
\midrule
\multicolumn{3}{l}{\textit{Unified Models}} \\
\rowcolor{lightblue}
Dynin-Robotics & 9.221 & 1.00 \\
\rowcolor{lightblue}
Dynin-Robotics-dInfer-BL7 & 91.236 $(9.89\times)$ & 6.81 \\
\rowcolor{lightblue}
Dynin-Robotics-dInfer-BL35 & \underline{268.834} $(29.15\times)$ & 20.23 \\
\bottomrule
\end{tabular}
\end{table}

\begin{table}[!htbp]
\centering
\caption{Block-parallel decoding throughput and reconstruction fidelity on LIBERO-Goal. All rows use the common one-B200 profiling protocol; parentheses report speedup over the Base decoder. Reconstruction metrics use held-out action chunks, and the best value in each metric is underlined.}
\label{tab:dinfer_acceleration}
\footnotesize
\setlength{\tabcolsep}{14pt}
\renewcommand{\arraystretch}{0.95}
\begin{tabular}{lccc}
\toprule
Configuration & Effective TPS~$\uparrow$ &
Token Acc (\%)~$\uparrow$ & Action MAE~$\downarrow$ \\
\midrule
Base & 9.221 & 99.7545 & \underline{0.021317} \\
\rowcolor{lightblue}
dInfer-BL7 & 91.236 $(9.89\times)$ & \underline{99.8214} & 0.021383 \\
dInfer-BL35 & \underline{268.834} $(29.15\times)$ & 99.4643 & 0.021468 \\
\bottomrule
\end{tabular}
\end{table}

The cross-model profiling results in Table~\ref{tab:throughput} provide context for the internal acceleration, while Table~\ref{tab:dinfer_acceleration} compares decoder variants on the same Dynin-Robotics model. BL7 increases throughput from 9.221 to 91.236 effective TPS, a $9.89\times$ speedup. BL35 reaches 268.834 TPS, or $29.15\times$ the base throughput. Thus, increasing the configured block size together with the optimized execution path substantially reduces the cost of producing an action chunk.

The reconstruction metrics show small changes in output fidelity across the decoder variants. BL7 has a token accuracy of 99.8214\%, compared with 99.7545\% for the base decoder, while its action MAE is slightly higher (0.021383 versus 0.021317). BL35 lowers token accuracy to 99.4643\% and raises MAE to 0.021468. Section~\ref{sec:decoding_sensitivity} complements these offline measurements with query latency, denoising work, and LIBERO-Goal rollout success.

\subsection{Ablation Study}
\label{sec:ablation}

The ablations examine three interventions: changing the initialization before Stage~2, changing the Stage-2 objective mixture, and changing inference composition after training. The training comparisons use the same reported 70k-step VLABench Stage-2 budget. In the objective-mixture comparison, inference variant~(e) is fixed while supervision changes; in the inference comparison, the checkpoint is fixed while the decoding composition changes. This design evaluates training recipes under a common decoder and inference choices with a common model. We report Track~1 success as ID, the average of Tracks~3 and~4 as OOD, and $(\mathrm{ID}-\mathrm{OOD})$ as the gap.

\subsubsection{Continual-Pretraining Ablation}

\begin{table}[!htbp]
\centering
% \caption{Effect of Stage-1 OXE continual pretraining under the same 70k-step VLABench Stage-2 protocol. The variants differ only in initialization from the 220.8k-micro-step OXE checkpoint. The best value in each metric is underlined.}
\caption{Effect of OXE continual pretraining on VLABench adaptation with a fixed 70k-step Stage-2 budget. Both variants build on Dynin-Omni; Stage~2 starts either directly from Dynin-Omni or from the checkpoint obtained after 220.8k OXE training micro-steps. ID denotes Track~1 success, and OOD is the mean success across Tracks~3 and~4. The higher success rate in each column is underlined.}
\label{tab:pretraining_ablation}
\footnotesize
\setlength{\tabcolsep}{10pt}
\renewcommand{\arraystretch}{0.95}
\begin{tabular}{lcccc}
\toprule
Training variant & Stage~1 & Stage~2 &
ID (\%)~$\uparrow$ & OOD (\%)~$\uparrow$ \\
\midrule
Without pretraining & \xmark & \cmark & 3.26 & 0.70 \\
\rowcolor{lightblue}
Dynin-Robotics & \cmark & \cmark & \underline{49.61} & \underline{47.28} \\
\bottomrule
\end{tabular}
\end{table}

Robot continual pretraining substantially improves adaptation within the reported Stage-2 budget. In Table~\ref{tab:pretraining_ablation}, initialization from Dynin-Omni without OXE continual pretraining reaches 3.26\% ID and 0.70\% OOD success, compared with 49.61\% and 47.28\% after initialization from the 220.8k-micro-step OXE checkpoint. Both pipelines start from the same omnimodal foundation, and the comparison measures the benefit of the additional robot-pretraining stage for downstream adaptation.

The improvement spans direct and shifted instructions. Under this recipe, 70k Stage-2 steps from the Dynin-Omni initialization do not recover the performance of the robot-pretrained model. The result identifies robot continual pretraining as a major contributor to downstream success within this fixed adaptation budget.

\subsubsection{Training-Objective Ablation}

\begin{table*}[!htbp]
\centering
% \caption{Cumulative Stage-2 objective ablation under the fixed 70k-step VLABench protocol. All rows are evaluated using inference variant~(e), goal-guided joint action--next-state denoising. The exposure ratio gives objective-sampling frequency rather than loss weights; row differences are conditional increments because objectives are added cumulatively. The best value in each metric is underlined.}
\caption{Cumulative Stage-2 objective mixtures under the 70k-step VLABench adaptation protocol. All variants start from the same Stage-1 checkpoint and are evaluated with goal-guided joint action--next-state denoising, variant~(e). Exposure ratios specify the relative frequency of the listed objectives in the training schedule, separately from their loss weights. Each row changes the active objectives and their exposure allocation. ID denotes Track~1 success, OOD averages Tracks~3 and~4, and Gap is ID minus OOD in percentage points. Underlining marks the highest success rates and the smallest gap.}
\label{tab:objective_ablation}
\footnotesize
\setlength{\tabcolsep}{6pt}
\renewcommand{\arraystretch}{0.95}
\begin{tabular}{llcccc}
\toprule
Training variant & Objectives & Exposure ratio &
ID (\%)~$\uparrow$ & OOD (\%)~$\uparrow$ & Gap (\%p)~$\downarrow$ \\
\midrule
Policy Only & $\{o_{\mathrm{PO}}\}$ & 1.0 & 47.15 & 33.88 & 13.27 \\
$+$ World Modeling & $\{o_{\mathrm{PO}},o_{\mathrm{WM}}\}$ & $0.8:0.2$ & 46.01 & 39.54 & 6.47 \\
$+$ Task Understanding & $\{o_{\mathrm{PO}},o_{\mathrm{WM}},o_{\mathrm{TU}}\}$ & $0.75:0.2:0.05$ & 49.38 & 45.29 & 4.09 \\
\rowcolor{lightblue}
$+$ Goal-State Prediction (All) & $\{o_{\mathrm{PO}},o_{\mathrm{WM}},o_{\mathrm{TU}},o_{\mathrm{GP}}\}$ & $0.65:0.2:0.05:0.10$ & \underline{49.61} & \underline{47.28} & \underline{2.33} \\
\bottomrule
\end{tabular}
\end{table*}

The Stage-2 mixture primarily changes performance under shifted instructions. Table~\ref{tab:objective_ablation} compares cumulative objective sets after 70k Stage-2 steps, evaluating every row with goal-guided joint action--next-state denoising, variant~(e). Relative to Stage-2 Policy-only training, the full mixture increases ID success from 47.15\% to 49.61\% and OOD success from 33.88\% to 47.28\%. The resulting improvements are $+2.46$ and $+13.40$ percentage points, respectively, and the gap decreases from 13.27 to 2.33 points. Here, the smaller gap accompanies substantially higher OOD success rather than a loss of ID performance.

The cumulative increments are not uniform. Adding World Modeling increases OOD success by 5.66 points while reducing ID by 1.14 points. Adding Task Understanding then raises both scores, and adding Goal-State Prediction provides a further 1.99-point OOD increase with a smaller 0.23-point ID increase. These observations support the full mixture as a stronger Stage-2 recipe for the evaluated instruction shifts. The exposure schedule changes alongside the objective set, with Policy exposure falling from 1.0 to 0.65, so each increment reflects both the added targets and a reallocation of supervision.

All variants inherit the Stage-1 checkpoint and use a decoder that consumes both goal and next-state predictions. The objective mixture can therefore improve the predictions used for action generation as well as the action predictor itself. The observed gain reflects the combined effect of adapting these interfaces under the shared inference procedure.

\subsubsection{Inference-Strategy Ablation}
\label{sec:inference_ablation}

\begin{table*}[!htbp]
\centering
% \caption{Inference composition on VLABench using the same post-trained checkpoint. Variant~(a) is action-only, variant~(c) is the default goal-guided path, and variants~(d) and~(f) rerank action candidates with World Modeling. The best value in each metric is underlined.}
\caption{Inference compositions on VLABench using the same post-trained checkpoint. The variants allocate inference computation to action decoding, goal prediction, joint action--next-state refinement, and world-model-based candidate reranking. Variant~(c) is the default path used for the reported LIBERO evaluations. ID denotes Track~1 success, OOD averages Tracks~3 and~4, and Gap is ID minus OOD in percentage points. Effective TPS follows the model-side timing scope defined in Section~\ref{subsec:evaluation_acceleration}. Underlining marks the highest success rates and throughput and the smallest gap.}
\label{tab:inference_ablation}
\footnotesize
\setlength{\tabcolsep}{8pt}
\renewcommand{\arraystretch}{0.95}
\begin{tabular}{lcccc}
\toprule
Inference variant & ID (\%)~$\uparrow$ & OOD (\%)~$\uparrow$ &
Gap (\%p)~$\downarrow$ & Effective TPS~$\uparrow$ \\
\midrule
(a) Action-Only Policy & 45.8 & 41.4 & 4.4 & \underline{9.238} \\
(b) Action/World-Model Joint Denoising & 46.4 & 40.5 & 5.9 & 8.805 \\
(c) Goal-State-Guided Policy (default) & 45.6 & 38.2 & 7.4 & 9.208 \\
(d) World-Model Filtering & 46.3 & 42.0 & 4.3 & 4.904 \\
\rowcolor{lightblue}
(e) Goal-Guided Joint Action/World Denoising & \underline{49.6} & 47.3 & 2.3 & 8.709 \\
(f) Goal-Guided World-Model Filtering & 48.9 & \underline{47.7} & \underline{1.2} & 4.828 \\
\bottomrule
\end{tabular}
\end{table*}

Inference composition determines whether the checkpoint's auxiliary outputs improve policy success. Table~\ref{tab:inference_ablation} holds the post-trained checkpoint fixed. Action-only decoding~(a) obtains 45.8\% ID and 41.4\% OOD success. Goal-guided Policy~(c), the default path used for the reported LIBERO evaluations, obtains 45.6\% and 38.2\%; its OOD result is 3.2 percentage points below action-only decoding. Joint action--next-state denoising without goal guidance~(b) also has a lower OOD score, 40.5\%. Neither auxiliary path in isolation improves OOD success in this comparison.

Combining goal guidance with joint denoising~(e) increases ID and OOD success to 49.6\% and 47.3\%, gains of 3.8 and 5.9 percentage points over action-only decoding. Goal guidance thus has different effects in the two decoding settings: it reduces OOD success when added to Policy decoding but improves it when added to joint action--next-state denoising. This pattern supports composing the goal and next-state interfaces together for the tested instruction shifts. Although~(c) is used in the reported LIBERO benchmark protocol, the VLABench comparison favors~(e) and~(f) in absolute success.

The compositions allocate additional inference computation in different ways and yield distinct performance--cost trade-offs. Variant~(e) reduces effective throughput from 9.238 TPS for action-only decoding to 8.709 TPS, a 5.7\% decrease. Goal-guided world-model filtering~(f) achieves the highest reported OOD success, 47.7\%, at 4.828 TPS. Compared with~(e), this is a 0.4-point OOD increase, a 0.7-point ID decrease, and approximately 44.6\% lower throughput. These variants instantiate test-time scaling through additional visual prediction and candidate evaluation. The comparison shows that the return on additional computation depends on its allocation: the combined goal-guided joint decoder improves success at a modest throughput cost, while filtering incurs a larger cost for a smaller additional OOD gain.

\subsection{Hyperparameter Sensitivity}
\label{sec:sensitivity}

We examine three operational choices: the resolution of action discretization, the granularity of token commitment during denoising, and the relative strength of auxiliary losses. The action-bin and loss-scale comparisons use the 70k-step VLABench Stage-2 protocol, while decoding is evaluated on LIBERO-Goal. These studies identify operating points and characterize performance sensitivity within the tested settings.

\subsubsection{Action Token Binning}
\label{sec:bin_sensitivity}

\begin{table*}[!htbp]
\centering
\caption{Action-bin sensitivity under the 70k-step VLABench Stage-2 protocol. Only the active Stage-2 bin count $B_A$ changes; Action MAE uses normalized action space, and ID/OOD are success rates. The Stage-1 initialization uses 256 bins. The best value in each metric is underlined.}
\label{tab:action_bin_sensitivity}
\footnotesize
\setlength{\tabcolsep}{16pt}
\renewcommand{\arraystretch}{0.95}
\begin{tabular}{ccccc}
\toprule
$B_A$ & Action MAE~$\downarrow$ &
Validation Token Acc~$\uparrow$ & ID (\%)~$\uparrow$ & OOD (\%)~$\uparrow$ \\
\midrule
16 & 0.09 & 0.75 & 27.84 & 25.05 \\
\rowcolor{lightblue}
32 & \underline{0.02} & \underline{0.89} & \underline{49.61} & \underline{47.28} \\
64 & 0.05 & 0.58 & 32.90 & 29.86 \\
\bottomrule
\end{tabular}
\end{table*}

Finer action discretization does not monotonically improve downstream performance. In Table~\ref{tab:action_bin_sensitivity}, 32 bins yield the lowest normalized action MAE (0.02) and the highest rollout success: 49.61\% ID and 47.28\% OOD. Both 16 and 64 bins increase MAE and reduce rollout success. We therefore use 32 bins as the operating point for this VLABench Stage-2 comparison.

Continuous-action error and rollout success provide a common basis for comparing bin counts. Exact token accuracy uses a different number of categories at each resolution and therefore measures a different classification problem. The Stage-1 model uses 256 bins, so selecting a downstream bin count also changes the mapping between action-token indices and continuous values during adaptation. The sweep evaluates these discretization choices within the robot-pretrained adaptation pipeline, where the intermediate resolution of 32 bins yields the lowest normalized action error and highest rollout success.

\subsubsection{Denoising and Block-Parallel Decoding}
\label{sec:decoding_sensitivity}

The decoding study tests how reductions in sequential denoising work translate into latency and policy success. For the fixed 35-token chunk, BL7 configures blocks at the granularity of one 7-DoF action, while BL35 spans all five actions. The block length is a decoding setting; the measured number of backbone evaluations captures the realized refinement work.

\begin{table*}[!htbp]
\centering
\caption{Denoising and control trade-offs on LIBERO-Goal under the common one-B200 profiling protocol. Evals./chunk is the mean number of backbone evaluations; amortized latency assumes execution of all five predicted actions. Success is the mean $\pm$ standard deviation, and the best value in each metric is underlined.}
\label{tab:dinfer_control_tradeoff}
\footnotesize
\setlength{\tabcolsep}{10pt}
\renewcommand{\arraystretch}{0.95}
\begin{tabular}{lcccc}
\toprule
Configuration & Evals./chunk~$\downarrow$ &
Query (ms)~$\downarrow$ & Amort. (ms/action)~$\downarrow$ &
Success (\%)~$\uparrow$ \\
\midrule
Dynin-Robotics-base & 35.00 & 3,795.7 & 759.1 & \underline{$97.8\pm0.4$} \\
\rowcolor{lightblue}
Dynin-Robotics-dInfer-BL7 & 5.14 & 383.6 & 76.7 & $97.7\pm0.2$ \\
Dynin-Robotics-dInfer-BL35 & \underline{1.73} & \underline{130.2} & \underline{26.0} & $97.2\pm0.4$ \\
\bottomrule
\end{tabular}
\end{table*}

Table~\ref{tab:dinfer_control_tradeoff} connects the offline comparison in Table~\ref{tab:dinfer_acceleration} to LIBERO-Goal rollouts. BL7 reduces the mean backbone evaluations per chunk from 35.00 to 5.14 and query latency from 3,795.7 to 383.6~ms. Its mean success rate is 97.7\%, compared with 97.8\% for the base decoder. BL35 further reduces the means to 1.73 evaluations and 130.2~ms, with a success rate of 97.2\%. The observed cost reductions therefore accompany a 0.1-point success decrease for BL7 and a 0.6-point decrease for BL35. The table reports the variability of the success measurements alongside these means.

The latency figures distinguish production of an action chunk from responding to a new observation. Dividing query latency by five gives amortized costs of 76.7 and 26.0~ms per action for BL7 and BL35, assuming that all five actions are executed before replanning. With one-step replanning, each decision instead incurs the full chunk-generation latency. These measurements consequently support faster model-side action production, while the achievable feedback rate depends on how the decoder is integrated with sensing and execution. Among the tested settings, BL7 offers a smaller observed success change and BL35 a shorter query latency.

\subsubsection{Auxiliary-Objective Loss Scale}
\label{sec:loss_scale_sensitivity}

\begin{table*}[!htbp]
\centering
\caption{Auxiliary-loss-scale sensitivity under the 70k-step VLABench Stage-2 protocol. The Policy, World Modeling, Task Understanding, and Goal-State Prediction exposure ratio is fixed at $0.65:0.20:0.05:0.10$ with $\lambda_{\mathrm{PO}}=1$; $\alpha$ scales the three auxiliary losses jointly. The best value in each metric is underlined.}
\label{tab:auxiliary_loss_sensitivity}
\footnotesize
\setlength{\tabcolsep}{12pt}
\renewcommand{\arraystretch}{0.95}
\begin{tabular}{ccccc}
\toprule
$\alpha$ & $(\lambda_{\mathrm{PO}},\lambda_{\mathrm{WM}},\lambda_{\mathrm{TU}},\lambda_{\mathrm{GP}})$ &
ID (\%)~$\uparrow$ & OOD (\%)~$\uparrow$ & Gap (\%p)~$\downarrow$ \\
\midrule
$0.5\times$ & $(1,0.5,0.5,0.5)$ & 49.42 & 45.91 & 3.51 \\
\rowcolor{lightblue}
$1.0\times$ & $(1,1,1,1)$ & \underline{49.61} & \underline{47.28} & 2.33 \\
$2.0\times$ & $(1,2,2,2)$ & 38.05 & 36.70 & \underline{1.35} \\
\bottomrule
\end{tabular}
\end{table*}

The loss-scale comparison separates objective exposure from the strength of the corresponding updates. In Table~\ref{tab:auxiliary_loss_sensitivity}, the Policy, World Modeling, Task Understanding, and Goal-State Prediction exposure ratio remains $0.65:0.20:0.05:0.10$, Policy loss weight is fixed at one, and $\alpha$ scales the three auxiliary losses together. Equal loss weights ($\alpha=1$) yield the highest observed ID and OOD success, 49.61\% and 47.28\%. Halving the auxiliary weights gives similar ID success (49.42\%) and a lower OOD score (45.91\%), whereas doubling them reduces both scores substantially, to 38.05\% and 36.70\%.

The largest performance change occurs when auxiliary losses are upweighted to $\alpha=2$. Although this setting has the smallest ID--OOD gap, 1.35 points, it also has the lowest success in both conditions. Equal weighting provides the best observed balance of ID and OOD success among the tested settings, with a smaller performance difference from $\alpha=0.5$ than from $\alpha=2$. This pattern indicates that the training recipe is more sensitive to increasing auxiliary weights above one than to halving them.

\section{Conclusion}
\label{sec:conclusion}

We presented Dynin-Robotics, a unified vision-language-action model built on Dynin-Omni that expresses robot control, visual prediction, and trajectory understanding through a shared masked-diffusion formulation. Representing language, visual states, goals, and actions in a common trajectory sequence allows the model to learn complementary prediction tasks and reuse its visual predictions during action generation and selection. The shared interface connects multi-objective robot learning with inference procedures that allocate computation to goal prediction, joint refinement, and candidate evaluation.

On two VLABench tasks, robot continual pretraining improves adaptation within a fixed Stage-2 step budget, and the full objective mixture yields higher shifted-instruction success than Policy-only post-training under the same coupled decoder. With the checkpoint fixed, combining goal guidance with joint action--next-state denoising improves shifted-instruction success over action-only decoding, whereas either component alone yields lower shifted-instruction success than action-only decoding. The benefit of using visual predictions therefore depends on how they are composed during inference. Evaluations on LIBERO, zero-shot LIBERO-Plus, and a Franka Research~3 platform show competitive policy performance across the reported settings. An optimized block-parallel implementation also provides up to a $29.2\times$ speedup in model-side action decoding over the base implementation under the profiling protocol.

The evaluation covers adjacent-future action-conditioned World Modeling and terminal Goal-State Prediction. Frame copying remains stronger on the reported whole-image metrics for adjacent-future prediction, motivating direct evaluation of action-dependent changes and their relationship to control performance. Task Understanding is evaluated qualitatively, and the physical study covers one robot platform and workspace. Extending visual dynamics evaluation to longer horizons, quantifying trajectory-to-language accuracy, and studying additional embodiments and sensor modalities are natural next steps for the shared trajectory formulation.

\clearpage

\begingroup
\hbadness=10000
\bibliographystyle{unsrt}
\bibliography{main}
\endgroup

% \appendix
% \input{input/appendix.tex}

\end{document}